\documentclass[11pt]{article}
\usepackage[preprint]{acl} 
\usepackage{times}
\usepackage{latexsym}
\usepackage[T1]{fontenc}
\usepackage[utf8]{inputenc}
\usepackage{inconsolata}

\usepackage[dvipsnames,table]{xcolor}
\usepackage{microtype}
\usepackage{hyperref}
\usepackage{url}
\usepackage{booktabs}
\usepackage{float}
\usepackage{flafter}
\usepackage{placeins}
\usepackage{graphicx}
\usepackage{multirow}
\usepackage{array}
\usepackage{amsmath}
\usepackage{amssymb}
\usepackage{enumitem}
\usepackage{adjustbox}
\usepackage{CJKutf8}
\usepackage{lineno}
\usepackage{threeparttable}
\usepackage{tabularx}
\usepackage{longtable}
\usepackage{pdflscape}
\usepackage{ragged2e}
\usepackage{siunitx}

\makeatletter
\@ifundefined{bng}{

\def\bng{\bngx}

\font\bngx=bang10

\def\*#1*#2{o\null{#2}{#1}}

\def\d#1{\oalign{\smash{#1}\crcr\hidewidth{$\!$\rm.}\hidewidth}}

\def\sh#1{\setbox0=\hbox{#1}%
     \kern-.02em\copy0\kern-\wd0
     \kern.04em\copy0\kern-\wd0
     \kern-.02em\raise.0433em\box0 }}{}
\makeatother

\newcolumntype{L}[1]{>{\RaggedRight\arraybackslash}p{#1}}
\newcolumntype{C}[1]{>{\Centering\arraybackslash}p{#1}}
\newcolumntype{Y}{>{\RaggedRight\arraybackslash}X}

\definecolor{darkblue}{HTML}{000099}
\hypersetup{colorlinks=true, citecolor=darkblue, linkcolor=darkblue, urlcolor=darkblue}

\title{Cross-Lingual Steering for Figurative Language Generation}

\author{
Linfeng Liu$^{1}$ \quad
Tiffany Zhan$^{2}$ \quad
Louie Hong Yao$^{3}$ \quad
Saptarshi Ghosh$^{1}$ \quad
Tianyu Jiang$^{1}$ \\
$^{1}$Department of Computer Science, University of Cincinnati \\
$^{2}$School of Computer Science, Carnegie Mellon University \\
$^{3}$Independent Researcher \\
\texttt{\{liu2lf,ghosh2si\}@mail.uc.edu}, \\
\texttt{tzhan2@andrew.cmu.edu},
\texttt{lhyao731@gmail.com}, 
\texttt{tianyu.jiang@uc.edu} 
}

\begin{document}

\maketitle

\begin{abstract}
Cross-lingual activation steering has been studied mainly for language-agnostic properties such as factual knowledge and reasoning. Its transferability remains unclear for figurative language, whose expressions are often tied to a particular language. In this work, we study cross-lingual steering of figurative language: whether steering vectors learned in one language can affect generation in another. We first show that within a single language, steering effectively increases the ratio of the intended figure of speech, especially for metaphor and idiom. Then under the cross-lingual setting, across 408 comparisons spanning four multilingual models, six languages, and four figures of speech, steering increases the figurative ratio from the unsteered 12.1\% to 20.4\%. To understand the mechanism, we decompose each language's own steering vector into the mean of steering vectors from other languages and a residual, and show that this mean direction matches or outperforms the language's own vector while the residual steers worse. This suggests that, despite the highly language-dependent nature of figurative expressions, a partially shared direction exists in the representation space that enables figurative knowledge to transfer across languages.
\end{abstract}

\section{Introduction}

\begin{figure}[!t]
\centering
\includegraphics[
width=\columnwidth,
keepaspectratio
]{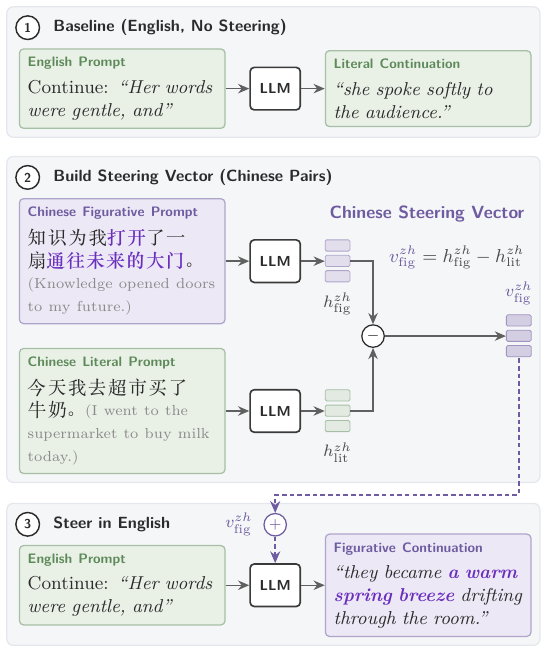}
\caption{Schematic overview of cross-lingual activation steering for figurative language generation. A steering direction estimated from Chinese metaphor examples and literal captions is applied to an English generation prompt. Chinese is the source language; English is the target language.}
\label{fig:crosslingual-steering-overview}
\end{figure}

Activation steering enables control of language model behavior at inference time. Studying whether this control transfers across languages can extend its use to multilingual settings and reveal shared representations across languages.
Studies of cross-lingual representations and model control have examined relatively language-agnostic properties, such as factual knowledge, reasoning, and safety~\citep{xu-etal-2023-language-representation,ifergan-etal-2025-beneath,maraia-etal-2026-activation,li2024preference}. 
In contrast, stylistic and rhetorical expressions often depend on language-specific conventions, yet steering studies of these phenomena remain largely English-focused~\citep{konen2024stylevectorssteeringgenerative,yao-etal-2026-rhetorical}.
This leaves open whether steering directions can transfer across languages when the intended behavior depends on language-specific forms of expression.

Figurative language provides a natural setting for studying cross-lingual transfer. Categories such as metaphor, idiom, irony, and sarcasm are found across languages. Their realization often depends on language-specific expressions, conceptual associations, and contextual cues. In this work, we investigate whether linear steering directions estimated for a figurative category in one language can induce the same category in text generated in another language. As an example, Figure~\ref{fig:crosslingual-steering-overview} shows how a metaphor direction estimated from Chinese examples is applied to English text generation.

To test whether figurative language generation can be steered across languages, we use an open-ended text generation task. We provide the model with a non-figurative caption from an image-captioning dataset and ask it to continue the text, without any instructions about writing style or figurative language. We then compare the generated text with and without steering, measuring how often it exhibits the intended figure of speech. We conduct this comparison in both \emph{monolingual steering}, where the direction is estimated in the generation language, and \emph{cross-lingual steering}, where it is estimated in another language. 
We further investigate whether steering directions share effective components across languages. For each target language, we test two directions: the average of same-category directions from the other languages and the residual obtained by removing the target-language direction's projection onto this average.

Across six languages and four multilingual large language models (LLMs), monolingual steering raises the proportion of outputs exhibiting the intended category above both unsteered generation and matched random-vector controls in all 24 model--language averages. Across 408 cross-lingual comparisons, the average gain over matched unsteered generation is 8.3 percentage points, with the largest gain for metaphor and the smallest for sarcasm. Comparing mean and residual steering directions connects cross-lingual alignment among the estimated directions to their effects on generation. Our contributions are:

\begin{itemize}

\item
\textbf{Monolingual figurative steering.}
We show that monolingual activation steering effectively controls figurative language generation across different models, languages, and figurative categories.

\vspace{-0.3em}

\item
\textbf{Cross-lingual transfer.}
Steering directions estimated in one language effectively increase the proportion of generated texts exhibiting the intended figurative category in another language, raising the average rate from 12.1\% to 20.4\% across 408 comparisons.
\vspace{-0.3em}

\item
\textbf{Shared steering components.}
We show that steering directions share effective components across languages. Mean directions achieve effective steering, while residual directions exhibit weaker or even no steering effects.

\end{itemize}

\section{Related Work}

\noindent \textbf{Activation steering for style and figurative language.}
Activation steering changes model behavior at inference time through interventions on internal activations~\citep{turner2024steeringlanguagemodelsactivation,zou2025representationengineeringtopdownapproach}. Contrastive Activation Addition (CAA) contrasts activations from examples with and without the desired property~\citep{rimsky-etal-2024-steering}, and related work has controlled properties including truthfulness and style~\citep{li2024inferencetimeinterventionelicitingtruthful,konen2024stylevectorssteeringgenerative}. Broader representation-engineering work systematizes these approaches~\citep{bartoszcze2025representationengineeringlargelanguagemodels}, while recent work probes and steers rhetorical properties~\citep{yao-etal-2026-rhetorical}. These studies remain English-focused, leaving activation steering for figurative language generation in other \mbox{languages underexplored}.

\noindent \textbf{Cross-lingual representations and steering.}
Multilingual models exhibit cross-language geometric and semantic alignment~\citep{chang-etal-2022-geometry,dumas-etal-2025-separating,brinkmann-etal-2025-large}. Linguistic profiling also relates embedding geometry to lexical and syntactic properties~\citep{domenichelli-etal-2026-linguistic}. Recent work shows that activation-based interventions can transfer across languages for reasoning and generation~\citep{maraia-etal-2026-activation,gurgurov2026clasbenchcrosslingualalignmentsteering,pokharel2026crosslingualactivationsteeringmultilingual}. Steering directions can be composed~\citep{scalena-etal-2024-multi}, and direction or subspace removal can test whether a representation contributes to model behavior~\citep{arditi2024refusal,shao2026decodeshare}. We examine alignment among independently estimated same-category directions by averaging directions from other languages and removing the target direction's projection onto their mean.

\noindent \textbf{Figurative language generation and style control.}
Research on figurative language has examined detection~\citep{CHEN2024127428}, interpretation~\citep{stowe-etal-2022-impli,chakrabarty-etal-2022-flute}, multilingual usage~\citep{park-etal-2025-fluid}, and cross-lingual metaphoricity~\citep{aghazadeh-etal-2022-metaphors}. Generation work includes reformulating text into a specified figure of speech~\citep{lai-nissim-2022-multi,10.1145/3654795} and generating contextually appropriate text after idiomatic or literal passages~\citep{pokharel-agrawal-2023-generating}. Activation-based work has also controlled stylistic properties through steering vectors and neuron interventions~\citep{konen2024stylevectorssteeringgenerative,lai-etal-2024-style}. Figurative categories combine lexical conventions, conceptual mappings, and pragmatic cues that can vary across languages. We test whether activation steering induces these categories under a generic generation instruction.

\section{Experimental Setup and Methodology}
\label{sec:methodology}

We evaluate steering through open-ended text generation, a setting used in prior activation-steering work~\citep{rimsky-etal-2024-steering}.

\noindent \textbf{Open-ended text generation.}
We study four figurative categories: idiom, metaphor, irony, and sarcasm. Idioms have conventional figurative meanings associated with the expression as a whole; metaphors describe one thing in terms of another. Irony contrasts intended meaning with literal wording, while sarcasm conveys criticism or mockery, often through irony. This range tests steering across lexical conventions, conceptual associations, and contextual interpretation.

Each main evaluation prompt pairs a held-out literal caption with a generic instruction to continue the text. Table~\ref{tab:generation-prompt-templates} lists the prompts; Appendix~\ref{appendix:generation} provides their Latin transliterations.

The \textit{source language} provides the figurative and literal examples used to estimate a steering direction, while the \textit{target language} is the language of generation. In monolingual steering, source and target are the same; in cross-lingual steering, they differ. Monolingual and cross-lingual steering use directions estimated for the intended \mbox{figurative category}.

\begin{table}[t]
\centering
\small
\renewcommand{\arraystretch}{1.18}
\begin{tabular}{@{}p{0.24\linewidth}p{0.68\linewidth}@{}}
\toprule
\textbf{Language} & \textbf{Generation prompt} \\
\midrule
English (en) & Continue the sentence: \{sent\} \\
Chinese (zh) & \begin{CJK*}{UTF8}{gbsn}续写这句话：\{sent\}\end{CJK*} \\
German (de) & Setze den Satz fort: \{sent\} \\
Spanish (es) & Continúa la oración: \{sent\} \\
Italian (it) & Continua la frase: \{sent\} \\
Bengali (bn) & {\bng bakY\*T*i\space{}ca\*l*i\*y*e\space{}Jan{\rm :}\bng \space{}{\rm \{sent\}}\bng } \\
\bottomrule
\end{tabular}
\caption{Language-specific instructions for open-ended text generation. In each template, \texttt{\{sent\}} is replaced with a held-out literal caption in the same language. The generic instruction leaves the figurative \mbox{category unspecified}.}
\label{tab:generation-prompt-templates}
\end{table}

\noindent \textbf{Experimental scope and data.}
We evaluate English, Chinese, Bengali, Spanish, Italian, and German using public resources with human-provided figurative annotations. The languages cover overlapping sets of figurative categories and differ in writing system and resource availability. The resulting data cover 22 language--category combinations across idiom, metaphor, irony, and sarcasm. The selected Spanish and German resources cover idiom, metaphor, and irony.

Direction estimation, layer validation, and final generation evaluation use disjoint data partitions. For each language--category combination, the direction-estimation set contains up to 500 human-annotated figurative examples and an equal number of same-language literal image captions. German metaphor and irony use 200 examples each because of smaller datasets. Final generation is evaluated on 500 held-out literal caption prompts per target language. Appendix~\ref{appendix:dataset-details} provides dataset sources, coverage, and sampling strategy.

\noindent \textbf{Steering direction estimation.}
We estimate steering directions using Contrastive Activation Addition (CAA)~\citep{rimsky-etal-2024-steering}. For language $g$ and category $c$, let $\mathcal{D}^{+}_{g,c}$ contain the figurative examples and $\mathcal{D}^{-}_{g}$ the same-language literal examples. Subtracting the mean literal activation from the mean figurative activation gives a vector pointing toward the figurative examples. At layer $l$, we normalize this difference to unit length:
\begin{equation}
\hat{v}_{g,c}^{(l)} =
\frac{
\mu^{(l)}(\mathcal{D}^{+}_{g,c}) -
\mu^{(l)}(\mathcal{D}^{-}_{g})
}{
\left\|
\mu^{(l)}(\mathcal{D}^{+}_{g,c}) -
\mu^{(l)}(\mathcal{D}^{-}_{g})
\right\|_2
}.
\end{equation}
Here, $\mu^{(l)}(\mathcal{D})$ averages the layer-$l$ activations at the final input token across all examples in $\mathcal{D}$.

\begin{table*}[!t]
\centering\small
\resizebox{\textwidth}{!}{%
\begin{tabular}{l ccc ccc ccc ccc}
\toprule
& \multicolumn{3}{c}{Qwen3-8B} & \multicolumn{3}{c}{Qwen3-32B} & \multicolumn{3}{c}{Llama-3.1-8B-Instruct} & \multicolumn{3}{c}{Ministral-3-8B-Instruct} \\
\cmidrule(lr){2-4}\cmidrule(lr){5-7}\cmidrule(lr){8-10}\cmidrule(lr){11-13}
Language & Unst. & Rand. & Steer & Unst. & Rand. & Steer & Unst. & Rand. & Steer & Unst. & Rand. & Steer \\
\midrule
English & 10.2 & 13.8 & 21.9 & 11.6 & 12.3 & 25.6 & 18.1 & 18.8 & 23.2 & 19.8 & 18.9 & 21.7 \\
Chinese & 14.7 & 15.0 & 28.0 & 14.6 & 15.0 & 27.3 & 7.8 & 9.1 & 10.2 & 22.5 & 24.2 & 29.7 \\
Bengali & 1.1 & 0.2 & 2.9 & 0.9 & 1.2 & 9.0 & 1.8 & 0.6 & 3.0 & 8.3 & 9.6 & 20.6 \\
Spanish & 5.4 & 4.8 & 12.9 & 5.9 & 5.6 & 15.1 & 11.1 & 10.3 & 15.6 & 17.2 & 14.6 & 20.7 \\
Italian & 7.6 & 8.2 & 23.2 & 8.3 & 8.8 & 27.4 & 11.8 & 10.7 & 17.1 & 21.9 & 22.5 & 30.9 \\
German & 7.8 & 7.6 & 20.2 & 8.9 & 8.6 & 28.3 & 9.0 & 9.1 & 16.1 & 24.4 & 25.1 & 32.5 \\
\bottomrule
\end{tabular}%
}
\caption{Monolingual steering's Target Behavior Rate (TBR, \%) by model and language, macro-averaged over the available study categories. \textsc{Unst.}, \textsc{Rand.}, and \textsc{Steer} denote unsteered generation, the matched random-vector control, and steering with the estimated direction, respectively.}
\label{tab:monolingual-language-summary}
\end{table*}

\begin{table}[t]
\centering
\small
\resizebox{\columnwidth}{!}{%
\begin{tabular}{l cc cc cc cc}
\toprule
& \multicolumn{2}{c}{Idiom}
& \multicolumn{2}{c}{Metaphor}
& \multicolumn{2}{c}{Sarcasm}
& \multicolumn{2}{c}{Irony} \\
\cmidrule(lr){2-3}
\cmidrule(lr){4-5}
\cmidrule(lr){6-7}
\cmidrule(lr){8-9}
Language & U & S & U & S & U & S & U & S \\
\midrule
English & 5.2 & 19.4 & 35.2 & 64.8 & 0.4 & 0.4 & 0.2 & 2.8 \\
Chinese & 6.0 & 12.8 & 52.4 & 89.6 & 0.0 & 3.8 & 0.2 & 5.8 \\
Bengali & 0.8 & 3.0 & 3.2 & 2.8 & 0.0 & 0.0 & 0.6 & 5.8 \\
Spanish & 0.4 & 7.0 & 15.8 & 30.8 & -- & -- & 0.0 & 0.8 \\
Italian & 0.8 & 3.6 & 28.2 & 84.0 & 0.4 & 1.0 & 1.0 & 4.0 \\
German & 6.2 & 26.6 & 16.2 & 31.2 & -- & -- &  1.0 & 2.8 \\
\bottomrule
\end{tabular}%
}
\caption{Qwen3-8B monolingual Target Behavior Rate (TBR, \%) by language and figurative category. U and S denote unsteered and steered generation, respectively. Dashes mark unavailable combinations.}
\label{tab:monolingual-detail-qwen}
\end{table}

\noindent \textbf{Models and intervention settings.}
We evaluate Qwen3-8B, Qwen3-32B~\citep{yang2025qwen3technicalreport}, Llama-3.1-8B-Instruct~\citep{grattafiori2024llama3herdmodels}, and Ministral-3-8B-Instruct~\citep{liu2026ministral3}. For each model, we select one intervention layer on the validation split and use it for the main steering experiments, with strength fixed at $\alpha=1.0$. Conditions use the same prompts and decoding settings. Random-vector controls replace the estimated direction with random directions matched in layer, token positions, and steering strength. Appendices~\ref{appendix:layer-selection} and~\ref{appendix:random-vector-control} detail the layer-selection procedure and the matched random-vector controls.

\noindent \textbf{Intervention application.}
During prompt processing, we add the normalized steering direction to the residual-stream activation at each prompt-token position $t$ in the selected layer:
\begin{equation}
h_{t}^{(l)\prime}=
h_{t}^{(l)}
+
\beta^{(l)}\hat{v}_{g,c}^{(l)}.
\end{equation}
We set $\beta^{(l)}=\alpha s^{(l)}$, where $s^{(l)}$ is a layer-specific reference activation norm (Appendix~\ref{appendix:generation-settings}) and $\alpha$ controls steering strength relative to that norm. We use $\alpha=1.0$ in the main experiments; Appendix~\ref{appendix:steering-strength} examines other strengths in a smaller English diagnostic. In monolingual steering and cross-lingual steering, $g$ is the source language. All generation conditions use target-language prompts.

\noindent \textbf{Evaluation and statistical analysis.}
We use DeepSeek-v4-flash~\citep{deepseekai2026deepseekv4} to judge whether each generated output exhibits the intended figurative category. Target Behavior Rate (TBR) is the proportion of outputs labeled as exhibiting the intended figure of speech. We separately score how coherently each output follows its prompt on a 0--4 scale. Appendix Tables~\ref{tab:deepseek-v4-flash-f1-yes_v1} and~\ref{tab:manual-audit-category-labels} report evaluator validation on annotated examples across all six languages and a manual audit of 800 English and Chinese Qwen3-8B monolingual outputs. The annotator was blinded to automatic labels. The audit yields 93.5\% agreement and 87.9\% F1 against the manual category labels.

We compare steered and unsteered outputs from the same prompts. Monolingual steering is also compared with random-vector controls. We report TBR differences in percentage points and use exact McNemar tests~\citep{mcnemar1947note}, with Benjamini--Hochberg (BH) correction for multiple comparisons~\citep{benjamini1995controlling}. Appendix~\ref{appendix:paired-statistics} details the pairing and correction families. Model--language summaries give equal weight to each available category.

Category judgments measure the presence of the intended figure of speech; coherence judgments measure how well the generated text follows its prompt. An output can exhibit the category while drifting from the input, or remain coherent while expressing literal content. We therefore report both measures and their joint occurrence.


\begin{table*}[t]
\centering\small
\resizebox{\textwidth}{!}{%
\begin{tabular}{l ccc ccc ccc ccc}
\toprule
& \multicolumn{3}{c}{Qwen3-8B} & \multicolumn{3}{c}{Qwen3-32B} & \multicolumn{3}{c}{Llama-3.1-8B-Instruct} & \multicolumn{3}{c}{Ministral-3-8B-Instruct} \\
\cmidrule(lr){2-4}\cmidrule(lr){5-7}\cmidrule(lr){8-10}\cmidrule(lr){11-13}
Language & Unst. & Mono. & Cross. & Unst. & Mono. & Cross. & Unst. & Mono. & Cross. & Unst. & Mono. & Cross. \\
\midrule
English & 10.2 & 21.9 & 26.0 & 11.6 & 25.6 & 26.2 & 18.1 & 23.2 & 22.2 & 19.8 & 21.7 & 22.5 \\
Chinese & 14.7 & 28.0 & 24.1 & 14.6 & 27.3 & 23.5 & 7.8 & 10.2 & 10.2 & 22.5 & 29.7 & 25.6 \\
Bengali & 1.1 & 2.9 & 0.6 & 0.9 & 9.0 & 2.6 & 1.8 & 3.0 & 2.0 & 8.3 & 20.6 & 13.3 \\
Spanish & 5.4 & 12.9 & 19.4 & 5.9 & 15.1 & 19.6 & 11.1 & 15.6 & 15.4 & 17.2 & 20.7 & 24.2 \\
Italian & 7.6 & 23.2 & 16.4 & 8.3 & 27.4 & 20.0 & 11.8 & 17.1 & 14.6 & 21.9 & 30.9 & 30.2 \\
German & 7.8 & 20.2 & 27.2 & 8.9 & 28.3 & 28.7 & 9.0 & 16.1 & 15.9 & 24.4 & 32.5 & 36.0 \\
\bottomrule
\end{tabular}%
}
\caption{Cross-lingual steering's Target Behavior Rate (TBR, \%) compared to unsteered generation and monolingual steering. Values are macro-averaged over the available study categories. \textsc{Mono.} uses the target-language direction. \textsc{Cross.} uses directions estimated in other languages separately and additionally averages TBR over the source languages available for each category.}
\label{tab:crosslingual-language-summary}
\end{table*}

\section{Monolingual Steering}
\label{sec:results-monolingual}

We first estimate and apply directions within each of the six study languages, establishing the monolingual reference for cross-lingual transfer. Unsteered generation provides the baseline TBR, and matched random-vector controls measure the effect of arbitrary perturbations at the selected layer and \mbox{steering strength}.

Table~\ref{tab:monolingual-language-summary} shows monolingual TBR across the study languages. Steering with the estimated directions yields higher TBR than both unsteered generation and the random-vector control in all 24 model--language averages, and significantly increases TBR over unsteered generation in 61 of 88 model--language--category combinations. Random-vector TBR generally stays close to the unsteered rate, supporting the contribution of the \mbox{estimated direction}.

Qwen3-32B shows large gains across most languages, whereas Llama shows more modest gains. Bengali is especially model-dependent: TBR stays close to the low unsteered rate under Qwen3-8B and Llama and increases more under Ministral.

The effect also varies by figurative category. Table~\ref{tab:monolingual-detail-qwen} shows positive TBR gains for Qwen3-8B in 19 of 22 language--category combinations. Idiom TBR increases in every language, metaphor shows several of the largest gains, irony shows more modest gains, and sarcasm shows the smallest gains. Across all four models, the proportion of model--language combinations with significant gains after BH correction is 88\% for metaphor, 83\% for idiom, 58\% for irony, and 38\% for sarcasm. Complete category-level results appear in Appendix~\ref{appendix:monolingual-detail}.

The absolute rates also differ across categories. Under Qwen3-8B, steered irony and sarcasm rates remain below 6\% in every evaluated language, whereas metaphor exceeds 60\% in English, Chinese, and Italian.

We also vary the literal examples used to estimate directions. The diagnostic in Appendix~\ref{appendix:matched-literal-mono} replaces image captions with literal examples from the figurative data sources. Both tested models retain positive average TBR gains, with gains varying across figurative categories.

\section{Cross-Lingual Transfer}
\label{sec:results-crosslingual}

We next test whether a steering direction remains effective when the generation language changes. The intended figurative category may require different lexical choices, semantic mappings, or pragmatic cues in the target language. We compare each cross-lingual intervention with matched unsteered generation. Throughout this section, \emph{gain} is the TBR difference from matched unsteered generation, measured in percentage points.

\begin{figure*}[t]
    \centering
    \includegraphics[width=0.9\textwidth]{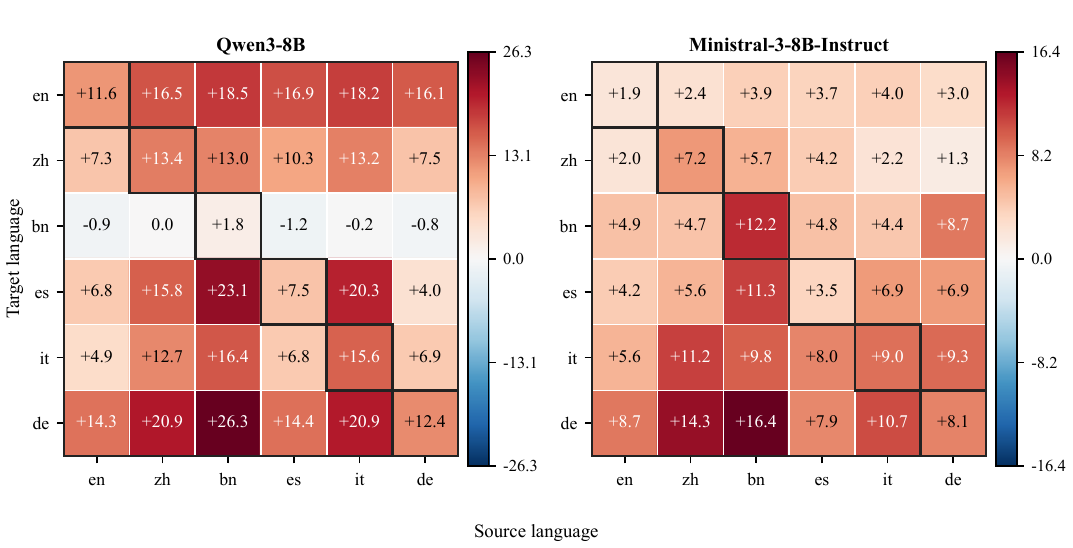}
    \caption{Steering gains by source and target language for Qwen3-8B and Ministral-3-8B-Instruct. Columns are source languages and rows are target languages. Off-diagonal cells show cross-lingual TBR gains over unsteered generation; outlined diagonal cells show the corresponding monolingual gains. Positive values indicate more generations exhibiting the intended category. Each panel uses its own symmetric color scale.}
    \label{fig:crosslingual-vs-mono-qwen8b-ministral}
\end{figure*}

\noindent \textbf{Overall cross-lingual transfer.}
Table~\ref{tab:crosslingual-language-summary} shows that cross-lingual steering raises TBR above unsteered generation in 23 of 24 model--language averages. We also compare it with monolingual steering, which uses a direction estimated directly in the target language. Cross-lingual TBR matches or exceeds monolingual TBR in many English, Spanish, and German settings, while it is consistently lower for Bengali and generally lower for Chinese and Italian. Directions can therefore remain effective across generation languages and sometimes match or exceed monolingual steering, depending on the target language and model.

Across all 408 comparisons spanning models, source and target languages, and figurative categories, cross-lingual steering raises mean TBR from 12.1\% under unsteered generation to 20.4\%, an average gain of 8.3 percentage points. After BH correction, 233 comparisons show significantly positive gains and 6 show significantly negative gains. Each of the four models has a positive average gain (Appendix Table~\ref{tab:crosslingual-category-model-summary}).

Each comparison fixes the model, source language, target language, and category and uses matched prompts. The overall cross-lingual gain gives equal weight to each of the 408 comparisons.

\noindent \textbf{Variation by source and target language.}
Figure~\ref{fig:crosslingual-vs-mono-qwen8b-ministral} shows source--target gains for Qwen3-8B and Ministral. Under Qwen3-8B, cross-lingual gains for Bengali targets are near zero, whereas gains for German targets range from 14.3 to 26.3 points and all exceed the 12.4-point monolingual gain. Under Ministral, every source language yields a positive average gain on the other target languages. German continues to show large gains, while Bengali gains 4.4--8.7 points from directions estimated in other languages. Thus, gains for the same target language vary across models, and different source languages can produce different gains for the same target. Corresponding matrices for Llama-3.1-8B-Instruct and Qwen3-32B appear in Appendix Figure~\ref{fig:crosslingual-vs-mono-llama8b-qwen32b}.

\begin{figure*}[t]
    \centering
    \includegraphics[width=\textwidth]{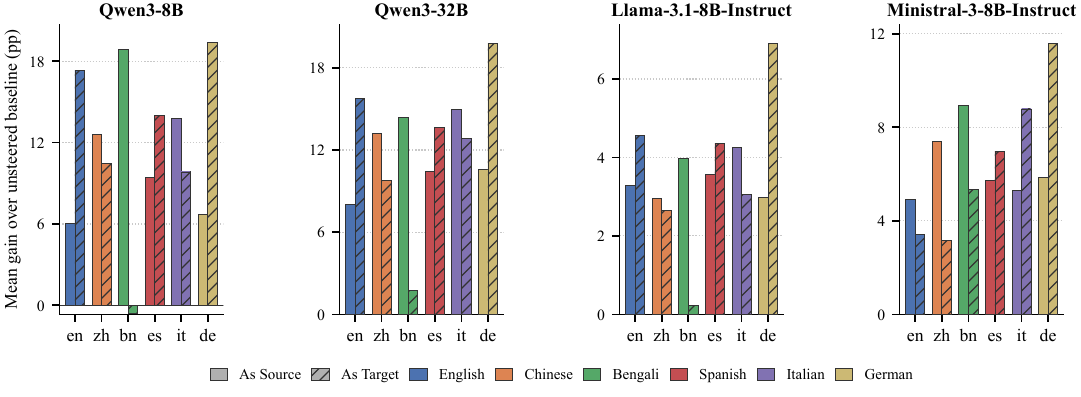}
    \caption{Average cross-lingual TBR gains by source and target language. For each language, ``As Source'' averages gains when its directions are applied in other languages, and ``As Target'' averages gains when directions from other languages are applied during generation in that language. Gains are measured relative to matched unsteered generation over the evaluated comparisons. Each model panel uses its own vertical scale.}
    \label{fig:macro-transfer-trajectories}
\end{figure*}

\noindent \textbf{Average gains as a source and as a target.}
These asymmetries distinguish a language's two roles in cross-lingual steering. Its \emph{average gain as a source} measures the TBR increase when its directions are applied in other languages, while its \emph{average gain as a target} measures the increase when directions from other languages are applied during generation in that language. Figure~\ref{fig:macro-transfer-trajectories} summarizes these two averages across all four models.

For Qwen3-8B, Bengali has gains of 18.9 points as a source and $-0.6$ points as a target. German shows the opposite pattern, with 6.7 points as a source and 19.4 points as a target. Across models, German consistently has one of the largest average gains as a target, while Bengali remains an effective source even when its average gain as a target is low. The larger Bengali target gains under Ministral show that target responsiveness also \mbox{depends on the model}.

Estimating transferable directions and responding to steering are therefore distinct aspects of multilingual generation. This distinction also appears when gains are averaged equally over idiom, metaphor, and irony, the three categories shared by all six languages (Appendix Table~\ref{tab:crosslingual-language-roles-balanced}).

Explicit prompting shows a similar model difference. On Bengali prompts, Qwen3-8B reaches 20.8\% TBR when instructed to produce the figurative category, while Ministral reaches 61.1\% (Appendix Table~\ref{tab:explicit-prompt-by-language}). This pattern is consistent with target gains partly reflecting how readily a model produces the intended category in that language.

\noindent \textbf{Variation by category and model.}
Transfer gains also vary substantially by figurative category. Appendix Table~\ref{tab:crosslingual-category-model-summary} shows the largest average gain for metaphor at 17.2 points, followed by 5.6 for idiom, 5.1 for irony, and 1.0 for sarcasm. Gains also vary across models: Qwen3-8B and Qwen3-32B show the largest average gains, Ministral shows a more moderate gain from a higher unsteered rate, and Llama shows the smallest gain.

For Qwen3-8B, we also replace literal captions in direction estimation with examples labeled as lacking the intended category, drawn from source-language figurative datasets. Average cross-lingual gains remain positive, with effects varying by figurative category (Appendix~\ref{appendix:alternative-negative-crosslingual}).

\begin{table*}[!t]
\centering
\begingroup
\setlength{\tabcolsep}{4pt}
\renewcommand{\arraystretch}{1.25}
\setlength{\emergencystretch}{1em}
\small
\begin{tabular}{@{}
    L{0.12\textwidth}
    L{0.21\textwidth}
    >{\fontsize{10}{11}\selectfont}L{0.29\textwidth}
    >{\fontsize{10}{11}\selectfont}L{0.29\textwidth}
@{}}
\toprule
\textbf{Model / category / source} &
\textbf{English input caption} &
\small\textbf{Generated text}\par\textbf{Unsteered} &
\small\textbf{Generated text}\par\textbf{Cross-lingual Steering} \\
\midrule
\textbf{Qwen3-8B}\par
{\textcolor{black!60}{\textsc{Metaphor}}}\par
{\textcolor{black!50}{\scriptsize DE$\rightarrow$EN}} &
\textit{A house interior with a plush chair, kitchen table and a view of the kitchen.} &
\ldots{} a cozy, well-lit space where natural light streams through large windows, with sleek countertops and a large \mbox{kitchen island}. &
\ldots{} Whether it is a quiet morning coffee or a lively dinner party, this kitchen becomes the \textbf{\emph{heart of the home}}. \\
\rowcolor{black!3}
\textbf{Llama-8B}\par
{\textcolor{black!60}{\textsc{Idiom}}}\par
{\textcolor{black!50}{\scriptsize ES$\rightarrow$EN}} &
\textit{A young boy in a blue uniform who just hit a baseball.} &
\ldots{} stood proudly at home plate, his eyes shining with excitement as the crowd erupted into cheers. &
\ldots{} was now running around the bases as he \textbf{\emph{grinned from ear to ear}}, the scoreboard behind him flashing ``HOME RUN.'' \\
\textbf{Ministral-8B}\par
{\textcolor{black!60}{\textsc{Irony}}}\par
{\textcolor{black!50}{\scriptsize ZH$\rightarrow$EN}} &
\textit{A woman talks on the phone while sitting at a table with a gold \mbox{coffee mug}.} &
``Excuse me, could you hold on just a second?'' \ldots{} She glances at the clock before \mbox{continuing the call}. &
``Oh, you're already on speaker? \textbf{\emph{Great}}. Now the whole office hears how much I \textbf{\emph{love}} this `urgent' thing he has `forgotten' about again.'' \\
\rowcolor{black!3}
\textbf{Qwen3-32B}\par
{\textcolor{black!60}{\textsc{Sarcasm}}}\par
{\textcolor{black!50}{\scriptsize IT$\rightarrow$EN}} &
\textit{A couple of guys with skateboards in the middle of the street.} &
\ldots{} were arguing over who had the right of way, oblivious to the honking cars and \mbox{growing crowd}. &
... in the middle of the street? Probably trying to grind the curb like it is a half-pipe \textbf{\emph{and the whole world isn't about to drive into them}}. \\
\bottomrule
\end{tabular}
\endgroup
\caption{Selected matched cross-lingual generation examples from four models and four source languages. Each direction is estimated in the source language shown in the first column and applied to English prompts. The unsteered output receives a negative target-category label, and the steered output receives a positive label. All displayed outputs receive a coherence score of 4. Bold italics mark the figurative expressions.}
\label{tab:crosslingual-examples}
\end{table*}

\noindent \textbf{Qualitative examples.}
Table~\ref{tab:crosslingual-examples} illustrates transfer through four directions applied to English generation prompts, spanning four source languages, categories, and models. The German-derived direction introduces an English metaphor, the Spanish-derived direction elicits a conventional English idiom, the Chinese-derived direction changes speaker stance through irony, and the Italian-derived direction produces sarcasm. The examples express the intended categories through English wording.

\begin{table*}[t]
\centering
\small
\resizebox{\textwidth}{!}{%
\begin{tabular}{c *{4}{S[table-format=2.1] S[table-format=2.1] S[table-format=2.1]}}
\toprule
 & \multicolumn{3}{c}{Qwen3-8B} & \multicolumn{3}{c}{Qwen3-32B} & \multicolumn{3}{c}{Llama-3.1-8B-Instruct} & \multicolumn{3}{c}{Ministral-3-8B-Instruct} \\
\cmidrule(lr){2-4}
\cmidrule(lr){5-7}
\cmidrule(lr){8-10}
\cmidrule(lr){11-13}
{Lang.} & {Mono.} & {OMean} & {Residual} & {Mono.} & {OMean} & {Residual} & {Mono.} & {OMean} & {Residual} & {Mono.} & {OMean} & {Residual} \\
\midrule
\textsc{en} & 21.9 & 29.8 & 8.6 & 25.6 & 31.5 & 12.4 & 23.2 & 24.3 & 15.8 & 21.7 & 23.3 & 18.4 \\
\textsc{zh} & 28.0 & 28.7 & 24.2 & 27.3 & 25.2 & 23.7 & 10.2 & 11.1 & 8.0 & 29.7 & 25.9 & 26.0 \\
\textsc{bn} & 2.9 & 0.2 & 1.5 & 9.0 & 3.4 & 3.2 & 3.0 & 2.2 & 2.4 & 20.6 & 14.4 & 17.2 \\
\textsc{es} & 12.9 & 23.7 & 3.5 & 15.1 & 24.7 & 4.7 & 15.6 & 17.2 & 8.9 & 20.7 & 25.6 & 15.3 \\
\textsc{it} & 23.2 & 16.7 & 19.5 & 27.4 & 23.2 & 18.1 & 17.1 & 15.7 & 15.2 & 30.9 & 31.2 & 27.5 \\
\textsc{de} & 20.2 & 27.7 & 11.4 & 28.3 & 34.7 & 12.9 & 16.1 & 18.4 & 12.1 & 32.5 & 37.8 & 26.3 \\
\bottomrule
\end{tabular}%
}
\caption{Target Behavior Rate (TBR, \%) by target language, macro-averaged over available categories as in Table~\ref{tab:crosslingual-language-summary}. \textsc{Mono.} uses the target-language direction; OMean uses the normalized mean of same-category directions from other languages; Residual uses the normalized target-language direction after removing its projection onto OMean.}
\label{tab:geometry-tbr-delta-vs-monolingual-four-category-geometry-only}
\end{table*}

We next examine whether alignment among directions estimated independently for the same category contributes to their steering effect.

\section{Cross-Lingual Alignment of Steering Directions}
\label{sec:internal-geometry}

We examine this alignment with two complementary operations on directions estimated independently for the same figurative category. Prior work has studied cross-lingual steering, direction composition, and direction removal largely as separate operations~\citep{maraia-etal-2026-activation,gurgurov2026clasbenchcrosslingualalignmentsteering,scalena-etal-2024-multi,arditi2024refusal}. Here we average other-language directions and remove the target direction's aligned component separately for each model. We then compare the resulting directions on the same generation task.

\noindent \textbf{Mean and residual steering directions.}
Let $t=(g_t,c_t)$ denote a target language--category combination, and let $\hat{v}_{g,c_t}$ be the normalized direction for category $c_t$ estimated in language $g$. We take $\mathcal{G}^{-}_t$ to be the languages with available data for category $c_t$, excluding $g_t$. We average their same-category directions and normalize the result to obtain the \textbf{other-language mean direction (OMean)}:
\begin{equation}
\bar{v}^{\mathrm{O}}_t
=
\operatorname{norm}\!\left(
\sum_{g \in \mathcal{G}^{-}_t}
\hat{v}_{g,c_t}
\right).
\end{equation}
Here, $\operatorname{norm}$ rescales a vector to unit length, so summing and averaging yield the same normalized direction. OMean excludes the target-language direction and weights the other languages equally. Steering with OMean tests whether their mean direction increases the intended category's frequency in target-language generation.

Cross-lingual steering applies one source-language direction at a time and averages the resulting TBRs. OMean averages the directions first and applies the resulting vector in a single steering condition. Generation depends nonlinearly on the intervention, so the two operations can produce different outcomes.

We next remove the target-language direction's component along OMean. Subtracting its projection onto OMean leaves the perpendicular component. We normalize this remaining vector to define the \textbf{Residual} steering condition:
\begin{equation}
r_t
=
\operatorname{norm}\!\left(
\hat{v}_{g_t,c_t}
-
\left\langle
\hat{v}_{g_t,c_t},
\bar{v}^{\mathrm{O}}_t
\right\rangle
\bar{v}^{\mathrm{O}}_t
\right).
\end{equation}
We compare the original target-language direction with the normalized remaining direction after removing its component along OMean, using the same intervention strength. We apply OMean and Residual to target-language prompts and compare the generated outputs with those from \mbox{monolingual steering}.

Following previous setting, we normalize the monolingual, OMean, and Residual directions and use the same steering strength $\alpha$.

Table~\ref{tab:geometry-tbr-delta-vs-monolingual-four-category-geometry-only} reports TBR for monolingual steering, OMean, and Residual under the same category-averaging convention as Table~\ref{tab:crosslingual-language-summary}. OMean matches or exceeds monolingual TBR in 15 of 24 model--language averages, with strong results for English, Spanish, and German. Chinese, Bengali, and Italian show more variable outcomes. Appendix Table~\ref{tab:combined-interventions-all-models} compares these conditions with unsteered generation, matched random-vector controls, and cross-lingual steering.

We further examine the Residual direction, obtained by projecting out the component along OMean from the target-language direction. Residual yields lower TBR than monolingual steering in all 24 model--language averages, linking the OMean-aligned component to steering effectiveness in target-language generation. Across the 88 model--language--category combinations, Residual remains above unsteered generation in 52 and equal in 8 more (Appendix~\ref{appendix:geometry-cell-detail}).

Together, the OMean and Residual results show that shared components contribute to figurative steering and help explain its transfer from source to target languages.

To examine category matching, we compare OMean with CrossCat (Appendix~\ref{appendix:crosscat}). CrossCat averages directions from other figurative categories using the same source languages. It increases mean TBR over unsteered generation, with TBR differences from OMean varying by model and category.

\noindent\textbf{Output coherence.} We also measure joint success: the proportion of outputs that exhibit the intended category and receive a coherence score of at least 2, corresponding to text that is understandable and related to the prompt. Monolingual and cross-lingual steering increase joint success over unsteered generation for all four models, while reducing mean coherence. Detailed results appear in Appendix~\ref{appendix:output-quality}.

\section{Conclusion}
\label{sec:conclusion}

Our results show that activation steering can control figurative language generation both within and across languages. 
Steering directions estimated in one language can effectively increase the occurrence of intended figurative categories in another, despite differences in how languages express the same figurative category.
We further show that steering directions share effective components across languages, with mean directions achieving effective steering while residual directions exhibit weaker or even no effects. These findings suggest that language-dependent figurative expressions can be controlled, in part, through shared directions in the representation space.

\clearpage

\section*{Limitations}

\noindent \textbf{Coverage and scope.}
The study covers six languages, four figurative categories, and four multilingual models. Five of the six languages are Indo-European. Category availability and construction-set sizes vary across languages, with smaller sets for German metaphor and irony.

\noindent \textbf{Direction construction and alignment.}
The within-corpus literal and alternative-negative diagnostics retain positive average effects, with category-level differences across constructions. The synthetic meaning-matched metaphor pilot retains gains for English and Chinese targets (Appendix~\ref{appendix:meaning-matched-pilot}). The OMean and Residual experiments connect alignment among estimated directions to their effects on generation. The English/Chinese CrossCat diagnostic also finds positive average gains from means constructed from other categories (Appendix~\ref{appendix:crosscat}). These directions can reflect genre, register, wording, and meaning alongside the intended category.

\noindent \textbf{Generation and evaluation.}
Open-ended generation from captions evaluates figurative expression in short contexts. Broader discourse and conversational settings would extend the analysis of irony and sarcasm. Category evaluation is automated at scale, with a manual audit of 800 Qwen3-8B monolingual outputs in English and Chinese. The annotator was blinded to automatic labels. A separate automated evaluation assesses naturalness and contextual fit among category-positive outputs (Appendix~\ref{appendix:conditional-quality}). Language-specific human assessments would extend evaluation of cultural fit and pragmatic intent. The main experiments use one validation-selected layer per model and strength $\alpha=1.0$; the sensitivity diagnostic covers English metaphor and sarcasm with Qwen3-8B (Appendix~\ref{appendix:steering-strength}).

\section*{Ethical considerations}

Figurative meaning, particularly irony and sarcasm, can depend strongly on cultural and conversational context, so steering may produce expressions that are inappropriate, misleading, or easily misinterpreted across communities. Such control could also be used to express manipulation, harassment, or misinformation indirectly, making harmful intent less apparent from literal wording alone. These risks motivate careful evaluation of cultural fit and downstream use when deploying activation steering for figurative language generation.

\bibliography{custom}
\clearpage
\appendix

\section{Dataset details and sampling strategy}
\label{appendix:dataset-details}

Our experiments cover four figurative categories (idiom, metaphor, irony, and sarcasm) in six languages: English (en), Chinese (zh), Bengali (bn), Spanish (es), Italian (it), and German (de).

\subsection{Dataset sources}
\label{appendix:dataset-sources}

We prioritize public resources with human-provided figurative annotations. Table~\ref{tab:dataset-sources} lists figurative resources and same-language literal-caption sources for the main experiments.

\begin{table*}[!t]
\centering
\footnotesize
\renewcommand{\arraystretch}{1.10}
\setlength{\tabcolsep}{2.8pt}
\begin{threeparttable}
\begin{tabularx}{\textwidth}{@{}l >{\RaggedRight\arraybackslash}X >{\RaggedRight\arraybackslash}X >{\RaggedRight\arraybackslash}X >{\RaggedRight\arraybackslash}X >{\RaggedRight\arraybackslash}X@{}}
\toprule
Language & Idiom & Metaphor & Irony & Sarcasm & Literal captions \\
\midrule
English & ID10M~\citep{tedeschi-etal-2022-id10m} & LCC~\citep{mohler-etal-2016-introducing} & SemEval-2018 Task 3~\citep{van-hee-etal-2018-semeval} & FLUTE~\citep{chakrabarty-etal-2022-flute} & MS COCO~\citep{lin2015microsoftcococommonobjects} \\
Chinese & Chinese Idiom Paraphrasing~\citep{qiang-etal-2023-chinese-idiom} & CCL 2018\tnote{1} & FGVIrony~\citep{WEN2025104169} & Topic-oriented sarcasm~\citep{liang-etal-2022-mian} & COCO-CN~\citep{8630050} \\
Bengali & \citet{sakhawat2026wordsdontmeansay} & \citet{das2025can} & Bengali tweets irony~\citep{ghosh2020irony} & Ben-Sarc~\citep{Lora_Shahariar_Nazmin_Rahman_Rahman_Bhuiyan_Shah_2024} & BAN-Cap~\citep{khan-etal-2022-ban} \\
Spanish & Spanish--Galician idioms~\citep{montesinos-etal-2026-improving} & LCC~\citep{mohler-etal-2016-introducing} & IroSvA~\citep{ortegaBueno2019irosva} & --- & MS-COCO-ES\tnote{2} \\
Italian & MultiCoPIE \citep{sentsova-etal-2025-multicopie} & PROMENADE~\citep{iuss_neplab_figurative_archive_2025} & EVALITA 2018~\citep{caselli2018evalita} & EVALITA 2018~\citep{caselli2018evalita} & \citet{IJCOL:scaiella_et_al:2019} \\
German & \citet{stap2024-idioms} & VOLIMET~\citep{piccirilli-etal-2024-volimet} ($N=200$) & MultiPICo~\citep{multipico} ($N=200$) & --- & COCO Karpathy OPUS\tnote{3} \\
\bottomrule
\end{tabularx}
\begin{tablenotes}[flushleft]
\scriptsize
\item[1] \href{https://github.com/DUTIR-Emotion-Group/CCL2018-Chinese-Metaphor-Analysis}{CCL 2018 Chinese Metaphor Analysis repository.}
\item[2] \href{https://github.com/carlosGarciaHe/MS-COCO-ES}{MS-COCO-ES repository.}
\item[3] \href{https://huggingface.co/datasets/Jotschi/coco-karpathy-opus-de}{COCO Karpathy OPUS German dataset.}
\end{tablenotes}
\caption{Dataset coverage for the main direction-estimation experiments. $N=200$ marks the two German language--category combinations that use fewer than the default maximum of 500 figurative examples. Dashes mark unavailable combinations.}
\label{tab:dataset-sources}
\end{threeparttable}
\end{table*}

\subsection{Sampling strategy and data partitions}
\label{appendix:sampling-strategy}

We keep direction estimation, layer validation, and final generation evaluation disjoint. For direction estimation, each evaluated language--category combination contains a balanced set of figurative and literal examples. We use the available annotated examples from each source, up to 500 figurative examples and 500 same-language literal captions per combination. German metaphor and irony each use 200 examples. Spanish and German sarcasm lack suitable annotated public datasets in the selected resources, giving 22 evaluated language--category combinations. The direction-estimation partition is used only to estimate the corresponding \mbox{category direction}.

\section{Evaluation and experimental details}
\label{appendix:evaluation-layer-selection}

We evaluate generated outputs with an LLM-based category evaluator. For each intended category, the evaluator assigns a binary label indicating whether an output exhibits that category. Target Behavior Rate (TBR) is the fraction of outputs for which the category evaluator returns \texttt{YES} for the intended figurative category. Prompt--output coherence is evaluated separately on a 0--4 scale. Complete prompts, construction data, evaluator outputs, and detailed results are provided in this \href{https://github.com/cincynlp/Xlingual-Figurative-Language/blob/main/ARTIFACT_MAP.md}{\mbox{GitHub repository}}.

\subsection{Evaluator validation and manual audit}
\label{appendix:manual-audit}

Table~\ref{tab:deepseek-v4-flash-f1-yes_v1} reports evaluator F1 on annotated figurative-versus-literal examples. An annotator, blinded to automatic labels, reviewed 800 Qwen3-8B monolingual outputs: 100 per category in English and Chinese. English metaphor and sarcasm samples use an earlier run under the same configuration. Table~\ref{tab:manual-audit-category-labels} compares the human and automatic labels.

\begin{table}[!t]
\centering
\footnotesize
\renewcommand{\arraystretch}{1.08}
\begin{tabular*}{\columnwidth}{@{\extracolsep{\fill}}lcccc@{}}
\toprule
Language & Idiom & Metaphor & Sarcasm & Irony \\
\midrule
English & 94.2 & 89.5 & 99.5 & 95.3 \\
Chinese & 93.0 & 95.9 & 95.8 & 90.1 \\
Spanish & 95.4 & 88.3 & N/A & 95.8 \\
German & 99.0 & 86.0 & N/A & 94.1 \\
Italian & 98.5 & 95.3 & 90.1 & 97.5 \\
Bengali & 82.8 & 75.8 & 87.0 & 83.5 \\
\bottomrule
\end{tabular*}
\caption{DeepSeek-v4-flash F1 scores (percent) on annotated figurative-versus-literal examples for the four study categories.}
\label{tab:deepseek-v4-flash-f1-yes_v1}
\end{table}

\begin{table}[!t]
\centering
\footnotesize
\renewcommand{\arraystretch}{1.08}
\begin{tabular*}{\columnwidth}{@{\extracolsep{\fill}}lrrrrr@{}}
\toprule
Subset & $N$ & Agreement & Precision & Recall & F1 \\
\midrule
English & 400 & 92.5 & 91.5 & 79.6 & 85.1 \\
Chinese & 400 & 94.5 & 96.2 & 85.0 & 90.3 \\
\rowcolor{black!3} Overall & 800 & 93.5 & 94.0 & 82.5 & 87.9 \\
\bottomrule
\end{tabular*}
\caption{Agreement between manual and automatic target-category labels (percent) over the four study categories. Precision, recall, and F1 treat the manual judgment as the reference.}
\label{tab:manual-audit-category-labels}
\end{table}

\subsection{Evaluator settings and coherence rubric}
\label{appendix:detector-prompt-definitions}

All DeepSeek-v4-flash~\citep{deepseekai2026deepseekv4} judgments use greedy decoding (temperature $=0.0$) and a maximum length of 256 tokens. The category evaluator makes one independent binary decision for the requested category, while the coherence evaluator considers contextual alignment, internal logic, and structural flow. Table~\ref{tab:coherence-evaluator-prompt} records the coherence scale used in the paper.

\begin{table}[t]
\centering\footnotesize
\renewcommand{\arraystretch}{1.06}
\begin{tabularx}{\columnwidth}{@{}c >{\RaggedRight\arraybackslash}X@{}}
\toprule
Score & Coherence criterion \\
\midrule
4 & Fully coherent: directly addresses the prompt, is internally consistent, and flows logically. \\
3 & Mostly coherent: generally logical and on task, with only a minor coherence or instruction-following issue. \\
2 & Partially coherent: understandable and related to the prompt, with a clear logical, organizational, or drift problem. \\
1 & Mostly incoherent: contains some relevant content and is difficult to follow, poorly connected, or substantially contradictory. \\
0 & Incoherent or failing: fundamentally hard to follow, largely fails to address the prompt, or is too disorganized to recover the intended meaning. \\
\bottomrule
\end{tabularx}
\caption{Compact coherence rubric used in the paper.}
\label{tab:coherence-evaluator-prompt}
\end{table}

\subsection{Generation prompts}
\label{appendix:generation}

The instructions in Table~\ref{tab:generation-prompt-templates} share the English meaning ``Continue the sentence.'' Latin transliterations of the Chinese and Bengali instructions are \textit{X\`uxi\v{e} zh\`e j\`u hu\`a} and \textit{B\=akya\d{t}i c\=ali\.y{e} y\=an}, respectively.

\subsection{Generation settings}
\label{appendix:generation-settings}

We use stochastic decoding for all unsteered and steered generation conditions, with identical decoding parameters within each model. Qwen3 models use temperature $=0.7$, $\texttt{top\_p}=0.8$, and $\texttt{top\_k}=20$; Llama-3.1-8B-Instruct and Ministral-3-8B-Instruct use temperature $=0.7$, $\texttt{top\_p}=0.9$, and $\texttt{top\_k}=50$. All conditions use a maximum generation length of 1024 tokens. Experiments were run on eight NVIDIA RTX 6000 Ada Generation GPUs, with an estimated total compute cost of approximately 900 GPU-hours.

For the main caption experiments, we compute $s^{(l)}$ before generation as the mean L2 norm of unsteered layer-$l$ outputs at the final token of chat-formatted \mbox{target-language prompts}.

\subsection{Validation-based layer selection}
\label{appendix:layer-selection}

We select one intervention layer per model using only the validation split and keep that layer fixed for the main steering experiments. Candidate relative depths are 0.40, 0.47, and 0.55 at $\alpha=1.0$. Each relative depth is mapped to the nearest valid transformer layer using half-up rounding. If at least one candidate has aggregate mean coherence above 2.5, we restrict selection to candidates above that threshold. Otherwise, all candidates remain eligible. We rank eligible candidates by TBR, breaking ties by coherence. Table~\ref{tab:selected-intervention-layers} reports the \mbox{selected layers}.

\begin{table}[!t]
\centering
\footnotesize
\renewcommand{\arraystretch}{1.08}
\begin{tabular*}{\columnwidth}{@{\extracolsep{\fill}}lcc@{}}
\toprule
Model & Relative depth & Layer \\
\midrule
Qwen3-8B & 0.40 & 14 \\
Qwen3-32B & 0.55 & 35 \\
Llama-3.1-8B-Instruct & 0.47 & 15 \\
Ministral-3-8B-Instruct & 0.55 & 19 \\
\bottomrule
\end{tabular*}
\caption{Validation-selected transformer layers and relative depths for the main experiments.}
\label{tab:selected-intervention-layers}
\end{table}

\subsection{Random-vector control}
\label{appendix:random-vector-control}

The random-vector control measures the effect of arbitrary hidden-state perturbations at the same intervention location and steering strength. For every evaluated model--language--category combination, the random direction is normalized and applied at the selected layer and prompt-token positions, with the same steering strength $\alpha$ used for the estimated direction. We use three independently sampled random directions (seeds 0, 1, and 2) and report mean TBR across the three runs. Averaging these runs reduces dependence on any single \mbox{sampled direction}.

\subsection{Paired statistical comparisons}
\label{appendix:paired-statistics}

The main test design uses 500 held-out prompts per evaluated cell. A small number of comparisons contain 498 or 499 matched pairs. Conditions are aligned by the held-out input text when the identifier is available and unique, and otherwise by stored row order. For a comparison between conditions $A$ and $B$, we report the percentage-point TBR difference and use the paired binary category labels for an exact two-sided McNemar test~\citep{mcnemar1947note}. The test is determined by the discordant counts: prompts whose outputs change from a negative category label under $A$ to a positive label under $B$, and prompts whose labels change in the reverse direction.

For the primary cross-lingual analysis, Benjamini--Hochberg correction~\citep{benjamini1995controlling} is applied jointly to all 408 comparisons across models, source and target languages, and categories. For monolingual cell-level analyses, the correction family contains the 22 evaluated language--category cells within each model. These adjusted tests determine the significance counts reported in the main text and Appendix Table~\ref{tab:crosslingual-category-model-summary}.

\section{Detailed monolingual steering results}
\label{appendix:monolingual-detail}

Table~\ref{tab:monolingual-detail-deltas} reports monolingual TBR gains by model, language, and category. Across the 88 evaluated cells, 61 gains are significant after within-model Benjamini--Hochberg correction. By category, the counts are 21/24 for metaphor, 20/24 for idiom, 14/24 for irony, and 6/16 for sarcasm.

\begin{table*}[!t]
\centering
\footnotesize
\renewcommand{\arraystretch}{1.08}
\setlength{\tabcolsep}{3.4pt}

\begin{minipage}[t]{0.485\textwidth}
\centering
\textbf{Qwen3-8B}\par\vspace{2pt}
\begin{tabular*}{\linewidth}{@{\extracolsep{\fill}}lrrrr@{}}
\toprule
Language & Idiom & Metaphor & Irony & Sarcasm \\
\midrule
English & +14.2 & +29.6 & +2.6 & 0.0 \\
Chinese & +6.8 & +37.2 & +5.6 & +3.8 \\
Bengali & +2.2 & -0.4 & +5.2 & 0.0 \\
Spanish & +6.6 & +15.0 & +0.8 & --- \\
Italian & +2.8 & +55.8 & +3.0 & +0.6 \\
German & +20.4 & +15.0 & +1.8 & --- \\
\bottomrule
\end{tabular*}
\end{minipage}\hfill
\begin{minipage}[t]{0.485\textwidth}
\centering
\textbf{Qwen3-32B}\par\vspace{2pt}
\begin{tabular*}{\linewidth}{@{\extracolsep{\fill}}lrrrr@{}}
\toprule
Language & Idiom & Metaphor & Irony & Sarcasm \\
\midrule
English & +12.8 & +25.0 & +10.4 & +7.6 \\
Chinese & +11.2 & +35.6 & +1.8 & +2.0 \\
Bengali & +2.0 & +7.6 & +22.8 & -0.2 \\
Spanish & +5.6 & +9.6 & +12.4 & --- \\
Italian & +4.6 & +44.6 & +17.0 & +9.8 \\
German & +23.8 & +5.0 & +29.4 & --- \\
\bottomrule
\end{tabular*}
\end{minipage}

\vspace{0.65em}
\begin{minipage}[t]{0.485\textwidth}
\centering
\textbf{Llama-3.1-8B-Instruct}\par\vspace{2pt}
\begin{tabular*}{\linewidth}{@{\extracolsep{\fill}}lrrrr@{}}
\toprule
Language & Idiom & Metaphor & Irony & Sarcasm \\
\midrule
English & +6.4 & +9.8 & +1.4 & +2.6 \\
Chinese & +3.2 & +5.6 & -0.4 & +1.2 \\
Bengali & +1.4 & +2.8 & -0.8 & +1.2 \\
Spanish & +5.8 & +7.0 & +0.8 & --- \\
Italian & +3.4 & +14.0 & +3.6 & +0.2 \\
German & +9.4 & +9.8 & +2.2 & --- \\
\bottomrule
\end{tabular*}
\end{minipage}\hfill
\begin{minipage}[t]{0.485\textwidth}
\centering
\textbf{Ministral-3-8B-Instruct}\par\vspace{2pt}
\begin{tabular*}{\linewidth}{@{\extracolsep{\fill}}lrrrr@{}}
\toprule
Language & Idiom & Metaphor & Irony & Sarcasm \\
\midrule
English & +0.6 & +8.0 & +1.4 & -2.2 \\
Chinese & +5.0 & +13.0 & +4.8 & +5.8 \\
Bengali & +6.0 & +38.2 & +1.2 & +3.4 \\
Spanish & -0.8 & +6.4 & +5.0 & --- \\
Italian & +3.8 & +18.0 & +10.8 & +3.6 \\
German & +7.6 & +14.2 & +2.6 & --- \\
\bottomrule
\end{tabular*}
\end{minipage}

\vspace{0.8em}
\begin{tabular*}{0.72\textwidth}{@{\extracolsep{\fill}}lrrr@{}}
\toprule
Category & Evaluated cells & BH-significant $+/-$ & Significant positive fraction \\
\midrule
Metaphor & 24 & 21/0 & 87.5\% \\
Idiom & 24 & 20/0 & 83.3\% \\
Irony & 24 & 14/0 & 58.3\% \\
Sarcasm & 16 & 6/0 & 37.5\% \\
\rowcolor{black!3} Overall & 88 & 61/0 & 69.3\% \\
\bottomrule
\end{tabular*}
\caption{Monolingual TBR gains (steered minus unsteered, percentage points) by model, language, and category. The summary reports statistically significant positive and negative comparisons after within-model BH correction. ``Significant positive fraction'' is the fraction of evaluated cells with a significantly positive gain. Dashes mark unavailable Spanish- and German-sarcasm cells.}
\label{tab:monolingual-detail-deltas}
\end{table*}

\section{Detailed cross-lingual transfer results}
\label{appendix:crosslingual-detail}

Table~\ref{tab:crosslingual-category-model-summary} summarizes the 408 primary cross-lingual comparisons across models, source and target languages, and categories.

\begin{table}[!t]
\centering
\footnotesize
\renewcommand{\arraystretch}{1.08}
\begin{tabular*}{\columnwidth}{@{\extracolsep{\fill}}lrrrrr@{}}
\toprule
Breakdown & $N$ & U & S & $\Delta$ & $+/-$ \\
\midrule
\rowcolor{black!3} Overall & 408 & 12.1 & 20.4 & +8.3 & 233/6 \\
\addlinespace[2pt]
\multicolumn{6}{l}{\textit{By category}} \\
Metaphor & 120 & 32.2 & 49.4 & +17.2 & 93/4 \\
Idiom & 120 & 5.8 & 11.4 & +5.6 & 72/0 \\
Irony & 120 & 2.6 & 7.7 & +5.1 & 58/0 \\
Sarcasm & 48 & 1.3 & 2.3 & +1.0 & 10/2 \\
\addlinespace[2pt]
\multicolumn{6}{l}{\textit{By model}} \\
Qwen3-8B & 102 & 8.5 & 20.0 & +11.4 & 70/4 \\
Qwen3-32B & 102 & 9.1 & 21.1 & +12.0 & 72/0 \\
Llama-8B & 102 & 10.6 & 14.1 & +3.5 & 36/0 \\
Ministral-8B & 102 & 20.1 & 26.4 & +6.4 & 55/2 \\
\bottomrule
\end{tabular*}
\caption{Cross-lingual transfer by figurative category and model. $N$ is the number of model-specific source--target--category comparisons. U and S are mean unsteered and steered TBR, and $\Delta$ is the mean paired gain in percentage points. $+/-$ counts significantly positive and negative comparisons after Benjamini--Hochberg correction over all 408 comparisons.}
\label{tab:crosslingual-category-model-summary}
\end{table}

\subsection{Average gains as a source and as a target}
\label{appendix:crosslingual-language-roles}

Table~\ref{tab:crosslingual-language-roles-balanced} recalculates average source and target gains using idiom, metaphor, and irony, the three categories shared by all six languages, with equal category weights. Bengali remains an effective source and shows small target gains in the Qwen and Llama models; German shows large target gains. English has the highest Qwen3-8B target gain under equal category weighting. The distinction between source and target effectiveness therefore also appears with common category coverage.

\begin{table*}[!t]
\centering
\footnotesize
\renewcommand{\arraystretch}{1.10}
\setlength{\tabcolsep}{3.2pt}
\begin{tabular*}{\textwidth}{@{\extracolsep{\fill}}l rr rr rr rr@{}}
\toprule
& \multicolumn{2}{c}{Qwen3-8B} & \multicolumn{2}{c}{Qwen3-32B} & \multicolumn{2}{c}{Llama-3.1-8B} & \multicolumn{2}{c}{Ministral-3-8B} \\
\cmidrule(lr){2-3}\cmidrule(lr){4-5}\cmidrule(lr){6-7}\cmidrule(lr){8-9}
Language & Src. & Tgt. & Src. & Tgt. & Src. & Tgt. & Src. & Tgt. \\
\midrule
English & +7.19 & +20.43 & +9.39 & +18.20 & +3.77 & +5.37 & +5.27 & +4.64 \\
Chinese & +14.80 & +12.39 & +15.67 & +11.49 & +3.41 & +3.04 & +8.61 & +3.40 \\
Bengali & +22.61 & -0.72 & +17.08 & +1.84 & +4.68 & +0.36 & +10.51 & +6.02 \\
Spanish & +9.45 & +13.99 & +10.44 & +13.65 & +3.56 & +4.36 & +5.71 & +6.99 \\
Italian & +16.47 & +11.81 & +17.08 & +15.28 & +5.13 & +3.49 & +6.37 & +9.70 \\
German & +6.73 & +19.36 & +10.59 & +19.77 & +2.97 & +6.91 & +5.86 & +11.59 \\
\bottomrule
\end{tabular*}
\caption{Average cross-lingual TBR gains as a source (Src.) and as a target (Tgt.) after balancing category coverage, in percentage points over unsteered generation. Values are restricted to idiom, metaphor, and irony, with equal weight for each category.}
\label{tab:crosslingual-language-roles-balanced}
\end{table*}

\begin{figure*}[t]
    \centering
    \includegraphics[width=0.94\textwidth]{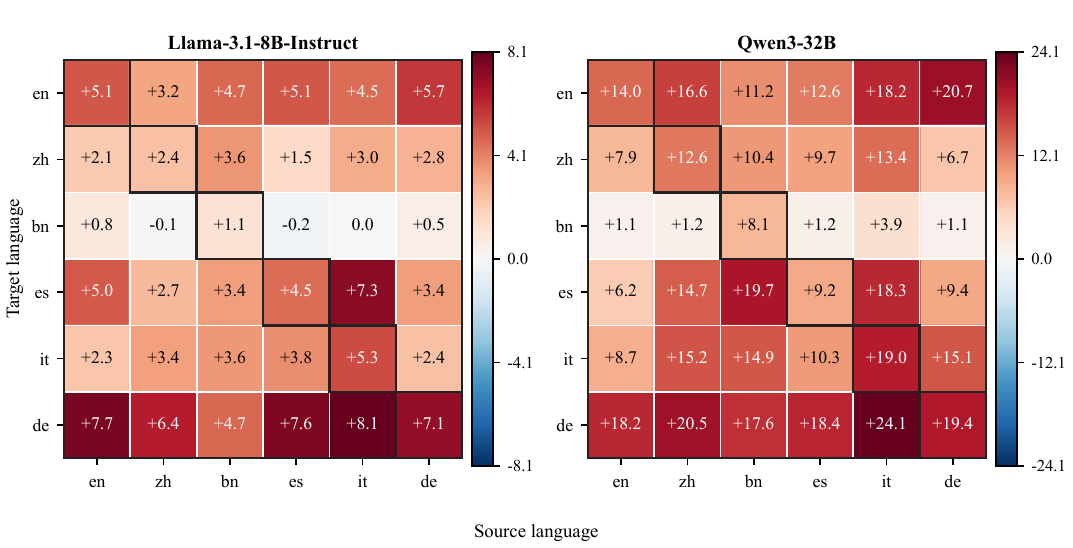}
    \caption{Steering effects across source and target languages for Llama-3.1-8B-Instruct and Qwen3-32B. Columns are source languages and rows are target languages. Off-diagonal cells show Cross-lingual TBR $-$ Unsteered TBR, while outlined diagonal cells show Monolingual TBR $-$ Unsteered TBR. Positive values indicate higher TBR under steering. Each panel uses its own symmetric color scale.}
    \label{fig:crosslingual-vs-mono-llama8b-qwen32b}
\end{figure*}

Figure~\ref{fig:crosslingual-vs-mono-llama8b-qwen32b} gives the source--target TBR-gain matrices for Llama-3.1-8B-Instruct and Qwen3-32B, complementing the Qwen3-8B and Ministral matrices in the main text. Llama shows smaller changes across most comparisons, with Bengali remaining close to the unsteered baseline as a target ($-0.2$ to $+0.8$ points from directions estimated in other languages). Qwen3-32B produces larger positive gains, although gains for Bengali targets remain smaller than for the other target languages, at \mbox{1.1--3.9 points}.

\subsection{Explicit-prompt reference by language}
\label{appendix:explicit-prompt-language}

Explicit prompting requests the figurative category directly, unlike the generic generation prompts used for steering. Table~\ref{tab:explicit-prompt-by-language} reports TBR and coherence by language for comparison with the gains observed when each language is used as a target.

\begin{table*}[!t]
\centering
\small
\renewcommand{\arraystretch}{1.08}
\begin{tabular*}{0.78\textwidth}{@{\extracolsep{\fill}}lrrrr@{}}
\toprule
& \multicolumn{2}{c}{Qwen3-8B} & \multicolumn{2}{c}{Ministral-3-8B} \\
\cmidrule(lr){2-3}\cmidrule(lr){4-5}
Language & TBR & Coh. & TBR & Coh. \\
\midrule
English & 62.5 & 3.72 & 91.8 & 3.76 \\
Chinese & 44.7 & 3.54 & 85.0 & 3.71 \\
Bengali & 20.8 & 2.88 & 61.1 & 3.16 \\
Spanish & 65.9 & 3.46 & 94.7 & 3.66 \\
Italian & 55.2 & 3.64 & 97.2 & 3.73 \\
German & 65.7 & 3.49 & 99.7 & 3.65 \\
\rowcolor{black!3} Overall & 51.2 & 3.45 & 87.4 & 3.61 \\
\bottomrule
\end{tabular*}
\caption{Explicit-prompt reference by prompt language. TBR is macro-averaged over the available four-category study cells for each language; coherence uses the 0--4 scale.}
\label{tab:explicit-prompt-by-language}
\end{table*}

\section{Detailed cross-lingual alignment results}
\label{appendix:combined-interventions}

We provide detailed comparisons for the mean and residual steering experiments in Section~\ref{sec:internal-geometry}. We first compare all steering conditions at the model--language level, then examine OMean and Residual gains over unsteered generation by language and category. Finally, CrossCat tests how these effects change when the source categories are mismatched. The differences and sign counts reported here are descriptive.

\subsection{Comparison across steering conditions}

Table~\ref{tab:combined-interventions-all-models} brings together unsteered generation, matched random-vector controls, monolingual steering, cross-lingual steering, OMean, and Residual. Values are macro-averaged over the available categories for each target language. Cross-lingual TBR averages the outcomes of separately applying directions from other languages, whereas OMean averages those directions before applying a single intervention. The monolingual, OMean, and Residual values follow Table~\ref{tab:geometry-tbr-delta-vs-monolingual-four-category-geometry-only}.

OMean has the highest TBR among the listed conditions for English, Spanish, and German under all four models. Chinese, Bengali, and Italian show more variable outcomes. Italian OMean TBR falls below monolingual TBR for both Qwen models and Llama and slightly exceeds it for Ministral. Residual falls below monolingual steering in all 24 model--language averages. Together, these comparisons connect the mean direction from other languages and the target direction's alignment with it to steering effectiveness.

\begin{table*}[!t]
\centering
\footnotesize
\renewcommand{\arraystretch}{1.08}
\setlength{\tabcolsep}{2.5pt}

\begin{minipage}[t]{0.485\textwidth}
\centering
\textbf{Qwen3-8B}\par\vspace{2pt}
\begin{tabular*}{\linewidth}{@{\extracolsep{\fill}}lrrrrrr@{}}
\toprule
Language & Unst. & Rand. & Mono. & Cross. & OMean & Res. \\
\midrule
English & 10.2 & 13.8 & 21.9 & 26.0 & 29.8 & 8.6 \\
Chinese & 14.7 & 15.0 & 28.0 & 24.1 & 28.7 & 24.2 \\
Bengali & 1.1 & 0.2 & 2.9 & 0.6 & 0.2 & 1.5 \\
Spanish & 5.4 & 4.8 & 12.9 & 19.4 & 23.7 & 3.5 \\
Italian & 7.6 & 8.2 & 23.2 & 16.4 & 16.7 & 19.5 \\
German & 7.8 & 7.6 & 20.2 & 27.2 & 27.7 & 11.4 \\
\bottomrule
\end{tabular*}
\end{minipage}\hfill
\begin{minipage}[t]{0.485\textwidth}
\centering
\textbf{Qwen3-32B}\par\vspace{2pt}
\begin{tabular*}{\linewidth}{@{\extracolsep{\fill}}lrrrrrr@{}}
\toprule
Language & Unst. & Rand. & Mono. & Cross. & OMean & Res. \\
\midrule
English & 11.6 & 12.3 & 25.6 & 26.2 & 31.5 & 12.4 \\
Chinese & 14.6 & 15.0 & 27.3 & 23.5 & 25.2 & 23.7 \\
Bengali & 0.9 & 1.2 & 9.0 & 2.6 & 3.4 & 3.2 \\
Spanish & 5.9 & 5.6 & 15.1 & 19.6 & 24.7 & 4.7 \\
Italian & 8.3 & 8.8 & 27.4 & 20.0 & 23.2 & 18.1 \\
German & 8.9 & 8.6 & 28.3 & 28.7 & 34.7 & 12.9 \\
\bottomrule
\end{tabular*}
\end{minipage}

\par\vspace{0.6em}
\begin{minipage}[t]{0.485\textwidth}
\centering
\textbf{Llama-3.1-8B-Instruct}\par\vspace{2pt}
\begin{tabular*}{\linewidth}{@{\extracolsep{\fill}}lrrrrrr@{}}
\toprule
Language & Unst. & Rand. & Mono. & Cross. & OMean & Res. \\
\midrule
English & 18.1 & 18.8 & 23.2 & 22.2 & 24.3 & 15.8 \\
Chinese & 7.8 & 9.1 & 10.2 & 10.2 & 11.1 & 8.0 \\
Bengali & 1.8 & 0.6 & 3.0 & 2.0 & 2.2 & 2.4 \\
Spanish & 11.1 & 10.3 & 15.6 & 15.4 & 17.2 & 8.9 \\
Italian & 11.8 & 10.7 & 17.1 & 14.6 & 15.7 & 15.2 \\
German & 9.0 & 9.1 & 16.1 & 15.9 & 18.4 & 12.1 \\
\bottomrule
\end{tabular*}
\end{minipage}\hfill
\begin{minipage}[t]{0.485\textwidth}
\centering
\textbf{Ministral-3-8B-Instruct}\par\vspace{2pt}
\begin{tabular*}{\linewidth}{@{\extracolsep{\fill}}lrrrrrr@{}}
\toprule
Language & Unst. & Rand. & Mono. & Cross. & OMean & Res. \\
\midrule
English & 19.8 & 18.9 & 21.7 & 22.5 & 23.3 & 18.4 \\
Chinese & 22.5 & 24.2 & 29.7 & 25.6 & 25.9 & 26.0 \\
Bengali & 8.3 & 9.6 & 20.6 & 13.3 & 14.4 & 17.2 \\
Spanish & 17.2 & 14.6 & 20.7 & 24.2 & 25.6 & 15.3 \\
Italian & 21.9 & 22.5 & 30.9 & 30.2 & 31.2 & 27.5 \\
German & 24.4 & 25.1 & 32.5 & 36.0 & 37.8 & 26.3 \\
\bottomrule
\end{tabular*}
\end{minipage}
\caption{Language-level Target Behavior Rate (TBR, \%) across the main steering conditions. Each panel reports one model. Res. denotes Residual steering.}
\label{tab:combined-interventions-all-models}
\end{table*}

\subsection{Results by language and category}
\label{appendix:geometry-cell-detail}

Language-level averages can conceal different effects across categories. Tables~\ref{tab:omean-vs-unsteered-cell-deltas} and~\ref{tab:residual-vs-unsteered-cell-deltas} report gains over unsteered generation for all 88 model--language--category cells. Positive entries indicate more outputs exhibiting the intended category. These tables show where each intervention retains an effect relative to unsteered generation; the comparison with monolingual steering above measures how its effectiveness changes after averaging or projection removal.

\subsubsection{OMean versus unsteered generation}

OMean raises TBR above unsteered generation in 77 of 88 cells, with 3 ties and 8 decreases (Table~\ref{tab:omean-vs-unsteered-cell-deltas}). Gains cover all 22 Qwen3-32B cells. Idiom and metaphor each show gains in 23 of 24 cells, with the decrease occurring for Bengali under Qwen3-8B. Irony and sarcasm show gains in 19 of 24 and 12 of 16 cells, respectively.

The magnitude of the gains also depends on the model and category. For Qwen3-8B, metaphor gains reach 53.8 points in English and 46.6 points in Spanish, while irony and sarcasm gains are much smaller. Qwen3-32B also shows large irony gains, including 30.8 points in English. Relative to monolingual steering, OMean is higher in 42 cells, equal in 4, and lower in 42, so its broad gains over unsteered generation coexist with mixed differences from monolingual steering.

\begin{table*}[!t]
\centering
\footnotesize
\renewcommand{\arraystretch}{1.10}
\setlength{\tabcolsep}{3.2pt}

\begin{minipage}[t]{0.485\textwidth}
\centering
\textbf{Qwen3-8B}\par\vspace{2pt}
\begin{tabular*}{\linewidth}{@{\extracolsep{\fill}}lrrrr@{}}
\toprule
Language & Idiom & Metaphor & Irony & Sarcasm \\
\midrule
English & +19.4 & +53.8 & +3.4 & +1.6 \\
Chinese & +13.4 & +39.6 & +2.2 & +1.0 \\
Bengali & -0.6 & -3.2 & 0.0 & 0.0 \\
Spanish & +6.6 & +46.6 & +1.6 & --- \\
Italian & +3.8 & +27.8 & +4.0 & +0.6 \\
German & +11.2 & +45.0 & +3.4 & --- \\
\bottomrule
\end{tabular*}
\end{minipage}\hfill
\begin{minipage}[t]{0.485\textwidth}
\centering
\textbf{Qwen3-32B}\par\vspace{2pt}
\begin{tabular*}{\linewidth}{@{\extracolsep{\fill}}lrrrr@{}}
\toprule
Language & Idiom & Metaphor & Irony & Sarcasm \\
\midrule
English & +8.6 & +37.2 & +30.8 & +2.8 \\
Chinese & +7.6 & +26.6 & +6.8 & +1.2 \\
Bengali & +0.4 & +3.2 & +4.0 & +2.2 \\
Spanish & +5.2 & +27.2 & +23.8 & --- \\
Italian & +6.6 & +23.8 & +25.8 & +3.0 \\
German & +19.6 & +29.2 & +28.6 & --- \\
\bottomrule
\end{tabular*}
\end{minipage}

\par\vspace{0.6em}
\begin{minipage}[t]{0.485\textwidth}
\centering
\textbf{Llama-3.1-8B-Instruct}\par\vspace{2pt}
\begin{tabular*}{\linewidth}{@{\extracolsep{\fill}}lrrrr@{}}
\toprule
Language & Idiom & Metaphor & Irony & Sarcasm \\
\midrule
English & +10.2 & +12.0 & +0.8 & +1.6 \\
Chinese & +3.4 & +9.2 & 0.0 & +0.6 \\
Bengali & +1.2 & +0.8 & -0.2 & -0.4 \\
Spanish & +7.0 & +10.6 & +0.8 & --- \\
Italian & +4.8 & +7.8 & +1.8 & +1.4 \\
German & +10.6 & +16.0 & +1.6 & --- \\
\bottomrule
\end{tabular*}
\end{minipage}\hfill
\begin{minipage}[t]{0.485\textwidth}
\centering
\textbf{Ministral-3-8B-Instruct}\par\vspace{2pt}
\begin{tabular*}{\linewidth}{@{\extracolsep{\fill}}lrrrr@{}}
\toprule
Language & Idiom & Metaphor & Irony & Sarcasm \\
\midrule
English & +5.8 & +12.8 & -0.8 & -3.8 \\
Chinese & +2.2 & +9.8 & -1.2 & +2.4 \\
Bengali & +3.4 & +20.7 & +0.2 & -0.4 \\
Spanish & +3.6 & +14.2 & +7.4 & --- \\
Italian & +2.0 & +12.7 & +16.2 & +6.6 \\
German & +18.6 & +20.2 & +1.2 & --- \\
\bottomrule
\end{tabular*}
\end{minipage}

\vspace{0.8em}
\begin{tabular*}{0.72\textwidth}{@{\extracolsep{\fill}}lrrr@{}}
\toprule
Model & Above unsteered & Equal & Below unsteered \\
\midrule
Qwen3-8B & 18 & 2 & 2 \\
Qwen3-32B & 22 & 0 & 0 \\
Llama-3.1-8B-Instruct & 19 & 1 & 2 \\
Ministral-3-8B-Instruct & 18 & 0 & 4 \\
\rowcolor{black!3} Overall & 77 & 3 & 8 \\
\bottomrule
\end{tabular*}
\caption{OMean TBR minus unsteered TBR (percentage points) for the 88 evaluated model--language--category cells. Signs indicate whether OMean is above or below the matched unsteered baseline; exact zeros indicate ties and dashes mark unavailable cells. The bottom panel reports the corresponding sign counts.}
\label{tab:omean-vs-unsteered-cell-deltas}
\end{table*}

\subsubsection{Residual versus unsteered generation}

\begin{table*}[!t]
\centering
\footnotesize
\renewcommand{\arraystretch}{1.10}
\setlength{\tabcolsep}{3.2pt}

\begin{minipage}[t]{0.485\textwidth}
\centering
\textbf{Qwen3-8B}\par\vspace{2pt}
\begin{tabular*}{\linewidth}{@{\extracolsep{\fill}}lrrrr@{}}
\toprule
Language & Idiom & Metaphor & Irony & Sarcasm \\
\midrule
English & +0.4 & -7.6 & 0.0 & +0.4 \\
Chinese & +5.0 & +30.4 & +1.2 & +1.4 \\
Bengali & 0.0 & -1.2 & +2.4 & 0.0 \\
Spanish & +2.6 & -8.8 & +0.4 & --- \\
Italian & +0.6 & +48.0 & -0.8 & -0.2 \\
German & +12.4 & -1.4 & -0.2 & --- \\
\bottomrule
\end{tabular*}
\end{minipage}\hfill
\begin{minipage}[t]{0.485\textwidth}
\centering
\textbf{Qwen3-32B}\par\vspace{2pt}
\begin{tabular*}{\linewidth}{@{\extracolsep{\fill}}lrrrr@{}}
\toprule
Language & Idiom & Metaphor & Irony & Sarcasm \\
\midrule
English & +2.6 & -1.8 & -0.4 & +2.4 \\
Chinese & +3.6 & +32.4 & -0.8 & +1.0 \\
Bengali & +0.2 & +8.4 & +0.6 & 0.0 \\
Spanish & +4.2 & -7.4 & -0.4 & --- \\
Italian & +1.8 & +37.0 & 0.0 & 0.0 \\
German & +14.0 & -4.8 & +2.8 & --- \\
\bottomrule
\end{tabular*}
\end{minipage}

\par\vspace{0.6em}
\begin{minipage}[t]{0.485\textwidth}
\centering
\textbf{Llama-3.1-8B-Instruct}\par\vspace{2pt}
\begin{tabular*}{\linewidth}{@{\extracolsep{\fill}}lrrrr@{}}
\toprule
Language & Idiom & Metaphor & Irony & Sarcasm \\
\midrule
English & -0.2 & -11.4 & -0.2 & +2.6 \\
Chinese & +1.0 & 0.0 & -0.6 & +0.4 \\
Bengali & 0.0 & +3.6 & -2.0 & +0.6 \\
Spanish & +0.8 & -7.6 & +0.2 & --- \\
Italian & +1.0 & +10.0 & +1.8 & +0.8 \\
German & +4.8 & +3.6 & +1.0 & --- \\
\bottomrule
\end{tabular*}
\end{minipage}\hfill
\begin{minipage}[t]{0.485\textwidth}
\centering
\textbf{Ministral-3-8B-Instruct}\par\vspace{2pt}
\begin{tabular*}{\linewidth}{@{\extracolsep{\fill}}lrrrr@{}}
\toprule
Language & Idiom & Metaphor & Irony & Sarcasm \\
\midrule
English & +1.0 & -3.0 & -1.2 & -2.2 \\
Chinese & -1.2 & +8.6 & +2.4 & +3.8 \\
Bengali & +2.6 & +30.1 & +0.6 & +2.1 \\
Spanish & -2.8 & -3.4 & +0.6 & --- \\
Italian & -0.6 & +18.3 & +3.4 & +1.4 \\
German & -3.6 & +9.6 & -0.4 & --- \\
\bottomrule
\end{tabular*}
\end{minipage}

\vspace{0.8em}
\begin{tabular*}{0.72\textwidth}{@{\extracolsep{\fill}}lrrr@{}}
\toprule
Model & Above unsteered & Equal & Below unsteered \\
\midrule
Qwen3-8B & 12 & 3 & 7 \\
Qwen3-32B & 13 & 3 & 6 \\
Llama-3.1-8B-Instruct & 14 & 2 & 6 \\
Ministral-3-8B-Instruct & 13 & 0 & 9 \\
\rowcolor{black!3} Overall & 52 & 8 & 28 \\
\bottomrule
\end{tabular*}
\caption{Residual TBR minus unsteered TBR (percentage points) for the 88 evaluated model--language--category cells. Signs indicate whether Residual is above or below the matched unsteered baseline. Exact zeros indicate ties, and dashes mark unavailable cells. The bottom panel reports the corresponding sign counts.}
\label{tab:residual-vs-unsteered-cell-deltas}
\end{table*}

Residual remains above unsteered generation in 52 cells, ties it in 8, and falls below it in 28 (Table~\ref{tab:residual-vs-unsteered-cell-deltas}). These remaining gains coexist with lower TBR than monolingual steering in every model--language average.

Metaphor shows a clear dependence on target language. Residual metaphor TBR falls below unsteered generation in English and Spanish and exceeds it in Italian for all four models. For Qwen3-8B, the corresponding gains are $-7.6$, $-8.8$, and $+48.0$ percentage points. Chinese metaphor also remains above unsteered generation for both Qwen models and Ministral, with a tie for Llama. The effect that remains after removing the mean direction therefore varies substantially across target languages, even within the same category.

\FloatBarrier
\subsection{Cross-category mean control}
\label{appendix:crosscat}

We test whether averaging directions from the intended category provides a benefit over averaging directions from other categories. \textbf{CrossCat} uses the same other-language source pool as OMean. Within each source language, we average the normalized directions for the available study categories other than the target category. We then average these per-language vectors with equal language weights and normalize the result to unit length. Both constructions exclude the target language; CrossCat also excludes the target category. All source vectors use caption negatives.

\begin{table}[t]
\centering
\footnotesize
\renewcommand{\arraystretch}{1.08}
\begin{tabular*}{\columnwidth}{@{\extracolsep{\fill}}lrrrr@{}}
\toprule
Model & Unsteered & OMean & CrossCat & $\Delta$ \\
\midrule
Qwen3-8B & 14.88 & 30.50 & 25.25 & +5.25 \\
Qwen3-32B & 16.00 & 28.63 & 23.63 & +5.00 \\
Llama-3.1-8B & 13.25 & 16.50 & 16.88 & -0.38 \\
Ministral-3-8B & 20.13 & 25.50 & 25.38 & +0.13 \\
\rowcolor{black!3} Overall & 16.06 & 25.28 & 22.78 & +2.50 \\
\bottomrule
\end{tabular*}
\caption{CrossCat control on English and Chinese targets. TBR (\%) averages eight language--category cells per model, with 100 matched test prompts per cell. Both means use the same source-language pool; CrossCat excludes the target category. $\Delta$ is OMean minus CrossCat in percentage points before rounding. Overall weights all 32 cells equally.}
\label{tab:crosscat-control}
\end{table}

We compare unsteered generation, OMean, and CrossCat for English and Chinese targets across all four models and study categories, using the first 100 held-out caption-test prompts per language. The layer, intervention strength, decoding settings, and norm calibration match each model's OMean run. We reuse the saved unsteered and OMean outputs and generate CrossCat outputs on the same prompts. All three conditions are evaluated afresh with DeepSeek-v4-flash through the official API in non-thinking mode, using the same target-category prompts. Each of the 32 model--language--category cells has 100 valid judgments per condition, for 9,600 judgments in total. Table~\ref{tab:crosscat-control} \mbox{reports mean TBR}.

Both means increase TBR over unsteered generation on average, by 9.22 percentage points for OMean and 6.72 for CrossCat. OMean's mean advantage is concentrated in the Qwen models, while Llama and Ministral have similar rates under both means. OMean has higher TBR in 17 of the 32 cells, equal TBR in 5, and lower TBR in 10. For Chinese metaphor, CrossCat exceeds OMean by 7--9 points in three of the four models.
\FloatBarrier

\section{Direction-construction diagnostics}
\label{appendix:construction-diagnostics}

The primary directions are estimated by contrasting figurative examples with literal captions. The first two diagnostics vary the negative examples. The final pilot constructs both figurative and literal examples as synthetic pairs intended to express the same meaning.

\subsection{Monolingual diagnostic with within-corpus literal examples}
\label{appendix:matched-literal-mono}

We replace literal captions as negative examples with literal examples from the same corpus for Llama-3.1-8B-Instruct and Qwen3-8B across the 22 available language--category combinations. Both constructions are evaluated on 100 held-out same-language WikiMatrix samples~\citep{schwenk2019wikimatrixmining135mparallel} using the primary intervention layer, $\alpha=1.0$, generation settings, and evaluator. Table~\ref{tab:wiki-input-caption-native-summary} summarizes the category-level gains.

\begin{table*}[!t]
\centering
\footnotesize
\renewcommand{\arraystretch}{1.10}
\begin{tabular*}{0.92\textwidth}{@{\extracolsep{\fill}}lrrc rrc@{}}
\toprule
& \multicolumn{3}{c}{Llama-3.1-8B-Instruct} & \multicolumn{3}{c}{Qwen3-8B} \\
\cmidrule(lr){2-4}\cmidrule(lr){5-7}
Category & Caption $\Delta$ & Corpus $\Delta$ & Sign retained & Caption $\Delta$ & Corpus $\Delta$ & Sign retained \\
\midrule
Idiom & +4.5 & +3.5 & 5/6 & +4.5 & +7.8 & 5/6 \\
Metaphor & +4.8 & +6.0 & 5/6 & +5.8 & -1.8 & 3/6 \\
Irony & -0.2 & +0.3 & 3/6 & +2.3 & +13.5 & 6/6 \\
Sarcasm & 0.0 & +0.3 & 4/4 & +0.3 & +0.3 & 4/4 \\
\rowcolor{black!3} Overall & +2.5 & +2.7 & 17/22 & +3.5 & +5.4 & 18/22 \\
\bottomrule
\end{tabular*}
\caption{Diagnostic with within-corpus literal examples. Entries are mean TBR gains in percentage points over the corresponding unsteered WikiMatrix baseline. ``Sign retained'' counts cells in which directions estimated with within-corpus literal examples preserve the non-negative versus negative gain obtained with caption negatives.}
\label{tab:wiki-input-caption-native-summary}
\end{table*}

Both sets of negative examples yield positive average effects, with differences across categories. For Qwen3-8B, literal examples from the same corpus increase the irony gain, while the mean metaphor gain changes from $+5.8$ to $-1.8$ points. The diagnostic shows that the choice of negative examples affects individual category gains.

\subsection{Cross-lingual alternative-negative diagnostic}
\label{appendix:alternative-negative-crosslingual}

In the second diagnostic, we estimate Qwen3-8B directions using non-target-category examples from the corresponding source-language figurative resources as the negative examples. We apply these directions to the same family of held-out caption prompts at the fixed layer and steering strength, using 50 prompts per comparison. Table~\ref{tab:native-negative-crosslingual} summarizes the 102 evaluated comparisons.

\begin{table}[!t]
\centering
\footnotesize
\renewcommand{\arraystretch}{1.08}
\begin{tabular*}{\columnwidth}{@{\extracolsep{\fill}}lrrrrr@{}}
\toprule
Category & $N$ & Unst. & Steer. & $\Delta$ & Positive \\
\midrule
Metaphor & 30 & 33.7 & 65.3 & +31.6 & 23 \\
Idiom & 30 & 5.0 & 15.4 & +10.4 & 21 \\
Irony & 30 & 1.0 & 3.9 & +2.9 & 13 \\
Sarcasm & 12 & 0.5 & 0.0 & -0.5 & 0 \\
\rowcolor{black!3} Overall & 102 & 11.7 & 24.9 & +13.1 & 57 \\
\bottomrule
\end{tabular*}
\caption{Qwen3-8B cross-lingual steering with non-target-category negative examples from the source-language figurative resources. $N$ counts source--target comparisons. Unst. and Steer. are mean TBR (\%); $\Delta$ is the mean gain in percentage points. The 102 comparison-level gains are macro-averaged. Positive counts comparisons with a positive TBR gain.}
\label{tab:native-negative-crosslingual}
\end{table}

\subsection{Meaning-matched synthetic metaphor pilot}
\label{appendix:meaning-matched-pilot}

We test whether steering remains effective when the figurative and literal examples used to estimate a direction express approximately the same meaning. DeepSeek-v4-flash generates English and Chinese metaphor--literal pairs, with instructions to preserve entities, topic, event, sentiment, tense, and speaker perspective where possible. It is also instructed to avoid duplicates, formulaic wording, and explicit simile markers. A separate automatic evaluation checks that each pair contains a metaphor and a literal reformulation with similar meaning, and screens both sentences for grammaticality, naturalness, and formulaic wording. We remove duplicates and randomly select 50 pairs per language from those accepted by the automatic evaluator.

For example, an English pair is ``His words were daggers to her heart'' and ``His words deeply hurt her emotionally.'' A Chinese pair, shown in English translation, is ``His words are knives'' and ``His words are hurtful''.

We estimate Qwen3-8B directions from these pairs and apply them at layer 14 with $\alpha=1.0$ in two monolingual and two cross-lingual comparisons. All conditions use the same 100 held-out literal caption prompts per target language from the validation split, with generation settings held fixed. We evaluate the outputs with DeepSeek-v4-flash through the official API in non-thinking mode, using the same category prompts across conditions. All 100 prompts per comparison have valid judgments under unsteered generation and both steering constructions. Table~\ref{tab:meaning-matched-pilot} reports the resulting TBRs.

\begin{table}[t]
\centering\footnotesize
\begin{tabular*}{\columnwidth}{@{\extracolsep{\fill}}llrrrr@{}}
\toprule
Source & Target & $N$ & Unst. & Caption & Paired \\
\midrule
English & English & 100 & 30.0 & 66.0 & 76.0 \\
English & Chinese & 100 & 49.0 & 77.0 & 83.0 \\
Chinese & English & 100 & 30.0 & 84.0 & 83.0 \\
Chinese & Chinese & 100 & 49.0 & 90.0 & 86.0 \\
\bottomrule
\end{tabular*}
\caption{Meaning-matched metaphor pilot with Qwen3-8B. Entries are TBR (\%). $N$ counts the evaluation prompts shared by all three conditions. Caption uses directions estimated from figurative examples and literal captions. Paired uses directions estimated from 50 synthetic figurative--literal pairs per source language.}
\label{tab:meaning-matched-pilot}
\end{table}

Directions estimated from synthetic pairs increase TBR over unsteered generation in all four comparisons: by 46.0 and 53.0 percentage points for English targets with English and Chinese source directions, respectively, and by 34.0 and 37.0 points for Chinese targets. Relative to caption-negative directions, paired directions yield higher TBR with English source directions (+10.0 points for English targets and +6.0 for Chinese targets) and slightly lower TBR with Chinese source directions (-1.0 and -4.0 points, respectively). Thus, meaning-matched construction preserves steering gains across both target languages in this pilot.

\section{Output coherence and figurative quality}
\label{appendix:output-quality}

TBR measures category occurrence; joint success additionally requires coherence. We compare both measures, examine steering strength, and assess quality among category-positive outputs.

\subsection{Joint success across steering conditions}

Joint success requires an output to exhibit the intended category and have a coherence score of at least 2. Table~\ref{tab:joint-success-ge2-main} compares unsteered generation, monolingual steering, cross-lingual steering, OMean, and Residual under this measure.

Monolingual and cross-lingual steering improve joint success over unsteered generation for every model, with cross-lingual steering remaining close to monolingual steering. OMean achieves the highest joint success for all four models, whereas Residual falls below OMean and monolingual steering and remains near or above the unsteered rate.

\begin{table}[!htbp]
\centering\small
\setlength{\tabcolsep}{4pt}
\resizebox{\columnwidth}{!}{%
\begin{tabular}{lrrrrr}
\toprule
Model & Unst. & Mono. & Cross. & OMean & Residual \\
\midrule
Qwen3-8B & 7.9 & 17.1 & 16.7 & 20.3 & 11.3 \\
Qwen3-32B & 8.4 & 19.8 & 18.4 & 22.9 & 12.8 \\
Llama-3.1-8B & 9.4 & 11.5 & 10.9 & 12.5 & 9.3 \\
Ministral-3-8B & 18.6 & 23.8 & 23.5 & 24.4 & 20.2 \\
\bottomrule
\end{tabular}%
}
\caption{Joint success (\%) across steering conditions. Success requires an output to exhibit the intended category and have a coherence score of at least 2. Cross-lingual comparisons are first averaged within each matched model--target-language--category cell. We then macro-average over the 22 study cells per model.}
\label{tab:joint-success-ge2-main}
\end{table}

\subsection{Coherence across steering conditions}

Table~\ref{tab:coherence-all-settings} reports mean coherence for the same intervention conditions. Unsteered generation has the highest mean coherence for every model. Monolingual and cross-lingual steering have similar coherence, with a maximum difference of about 0.06 points. For each model, Residual has higher mean coherence and lower average TBR and joint success than monolingual and cross-lingual steering.

\begin{table}[!htbp]
\centering
\footnotesize
\renewcommand{\arraystretch}{1.08}
\setlength{\tabcolsep}{3.2pt}
\begin{adjustbox}{max width=\columnwidth}
\begin{tabular}{@{}lccccc@{}}
\toprule
Model & Unst. & Mono. & Cross. & OMean & Resid. \\
\midrule
Qwen3-8B & 3.77 & 3.02 & 2.96 & 3.14 & 3.16 \\
Qwen3-32B & 3.94 & 3.40 & 3.43 & 3.58 & 3.68 \\
Llama-3.1-8B-Instruct & 3.36 & 2.88 & 2.86 & 2.89 & 3.07 \\
Ministral-3-8B-Instruct & 3.16 & 2.62 & 2.67 & 2.62 & 2.73 \\
\bottomrule
\end{tabular}
\end{adjustbox}
\caption{Mean coherence on the 0--4 scale across steering conditions. Cross-lingual comparisons are averaged within matched target-language--category cells before the model-level macro-average.}
\label{tab:coherence-all-settings}
\end{table}

\subsection{TBR and coherence under Qwen3-8B cross-lingual steering}
\label{appendix:crosslingual-quality}

Across Qwen3-8B's 102 cross-lingual comparisons (51,000 matched output pairs), steering increases TBR and joint success at a coherence threshold of at least 2, and also increases the proportion of outputs with coherence below 2.

\begin{table}[!htbp]
\centering
\footnotesize
\setlength{\tabcolsep}{5pt}
\renewcommand{\arraystretch}{1.08}
\begin{tabular*}{\columnwidth}{@{\extracolsep{\fill}}lrrr@{}}
\toprule
Measure & Unsteered & Cross-lingual & $\Delta$ (pp) \\
\midrule
Target Behavior Rate & 8.5 & 20.0 & +11.4 \\
Joint success & 8.5 & 17.9 & +9.4 \\
Low coherence & 1.5 & 20.9 & +19.4 \\
\bottomrule
\end{tabular*}
\caption{TBR and coherence for Qwen3-8B over 102 matched cross-lingual comparisons (51,000 output pairs). Joint success requires the intended category and coherence $\geq2$. Low coherence is a score below 2.}
\label{tab:crosslingual-joint-success}
\end{table}

\subsection{Steering-strength sweep}
\label{appendix:steering-strength}

We examine how TBR and coherence change with steering strength using Qwen3-8B and 100 held-out English caption prompts per category from the test split. The comparison covers metaphor and sarcasm. At layer 14, used in the main experiments, we test $\alpha\in\{0.5,1.0,1.5,2.0\}$. We also test layers 17 and 20 at $\alpha\in\{1.0,1.5\}$. Directions are estimated using caption negatives, with prompts and decoding settings held fixed. Unsteered TBR is 35.0\% for metaphor and 0.0\% for sarcasm, each with 100 valid category judgments.

\begin{table}[!htbp]
\centering\footnotesize
\setlength{\tabcolsep}{3pt}
\begin{tabular*}{\columnwidth}{@{\extracolsep{\fill}}rrrr@{}}
\toprule
$\alpha$ & TBR (valid $N$) & Coherence (valid $N$) & Low (\%) \\
\midrule
\multicolumn{4}{@{}l}{\textit{Metaphor}} \\
0.5 & 37.8 (74) & 3.98 (100) & 0.0 \\
1.0 & 65.0 (100) & 3.93 (100) & 0.0 \\
1.5 & 51.2 (86) & 3.38 (91) & 11.0 \\
2.0 & 45.5 (77) & 0.94 (77) & 70.1 \\
\midrule
\multicolumn{4}{@{}l}{\textit{Sarcasm}} \\
0.5 & 0.0 (100) & 4.00 (99) & 0.0 \\
1.0 & 1.0 (100) & 3.98 (98) & 0.0 \\
1.5 & 0.0 (99) & 0.76 (72) & 72.2 \\
2.0 & 0.0 (100) & 0.39 (70) & 82.9 \\
\bottomrule
\end{tabular*}
\caption{Steering strength at layer 14 for Qwen3-8B on 100 English prompts per category. TBR is a percentage, and coherence is the mean on the 0--4 scale. Parentheses give valid-judgment counts. Low is the percentage of valid coherence scores below 2.}
\label{tab:steering-strength-sweep}
\end{table}

At layer 14, metaphor TBR is highest at $\alpha=1.0$ among the tested strengths. Increasing $\alpha$ to 2.0 reduces mean coherence from 3.93 to 0.94 and raises the low-coherence rate to 70.1\%. Sarcasm TBR stays at 0--1\% at this layer, while low coherence reaches 72.2\% at $\alpha=1.5$ and 82.9\% at $\alpha=2.0$. Layer choice also affects metaphor generation: at $\alpha=1.0$, TBR is 65.0\%, 40.5\%, and 16.9\% at layers 14, 17, and 20. These English comparisons show how intervention strength and layer choice influence category occurrence and coherence.

For the eight layer-14 conditions, 736 of 800 outputs have valid category judgments and 707 have valid coherence scores. Each metric is computed over its valid judgments, with the corresponding counts shown in Table~\ref{tab:steering-strength-sweep}.

\subsection{Conditional figurative-quality evaluation}
\label{appendix:conditional-quality}

We evaluate naturalness, contextual fit, and the effectiveness of figurative use among Qwen3-8B outputs labeled as exhibiting the intended category. We compare unsteered generation, monolingual steering, and cross-lingual steering using the held-out caption-test outputs, with steering at layer 14 and $\alpha=1.0$. In a separate evaluation call, DeepSeek-v4-flash receives the target language, category and definition, original prompt, generated text, and a 0--4 quality rubric. It returns a quality score and a short reason at temperature 0, without access to the intervention condition, source language of the direction, coherence score, or earlier category judgment.

We evaluate all available category-positive outputs without a coherence threshold and average valid quality scores. Table~\ref{tab:conditional-quality-summary} gives equal weight to the 19 language--category cells with valid scores in all three conditions. Cross-lingual outputs are pooled across available source languages within each target-language--category cell before averaging. Confidence intervals use 10,000 paired bootstrap resamples of these cells.

\begin{table}[!htbp]
\centering\footnotesize
\setlength{\tabcolsep}{3pt}
\begin{tabular*}{\columnwidth}{@{\extracolsep{\fill}}lrrc@{}}
\toprule
Condition & $N$ & Quality & 95\% CI \\
\midrule
Unsteered & 864 & 2.50 & [1.98, 2.97] \\
Monolingual & 1,877 & 2.80 & [2.33, 3.19] \\
Cross-lingual & 9,965 & 2.67 & [2.20, 3.07] \\
\bottomrule
\end{tabular*}
\caption{Conditional figurative quality for Qwen3-8B, macro-averaged over the 19 language--category cells shared by all three conditions. $N$ counts valid quality scores in these cells. Quality and confidence intervals use the 0--4 scale.}
\label{tab:conditional-quality-summary}
\end{table}

The paired quality differences from unsteered generation are $+0.29$ (95\% CI $[-0.27,+0.83]$) for monolingual steering and $+0.17$ ($[-0.38,+0.71]$) for cross-lingual steering. Both confidence intervals include zero. These conditional comparisons can include different prompts in each condition, and the same model family supplies category and quality judgments.

\end{document}